\documentclass[conference]{IEEEtran}
\IEEEoverridecommandlockouts

\usepackage{amsmath,amsthm,amsbsy,amssymb,amsfonts,mathtools,multirow,soul,comment,hyperref,afterpage,wrapfig,booktabs,textcomp,tablefootnote,adjustbox,graphicx,wrapfig,float,stfloats,subcaption,cite,makecell}
\usepackage{algorithm,algorithmicx}
\usepackage[noend]{algpseudocode}
\usepackage[table]{xcolor}
\usepackage[T1]{fontenc}

\newtheoremstyle{spacedbolddef}
  {10pt}   % Space above
  {10pt}   % Space below
  {\normalfont}  % Body font
  {}       % Indent amount
  {\bfseries} % Theorem head font (bold)
  {.}      % Punctuation after theorem head
  {1em}    % Space after theorem head
  {}       % Theorem head spec

\theoremstyle{spacedbolddef}

\newcommand{\gray}[1]{\textcolor{gray}{#1}}

\begin{document}

\title{On the Fragility of Contribution Evaluation in Federated Learning}

\author{
    \IEEEauthorblockN{Balazs Pejo \& Gergely Biczok}
    \IEEEauthorblockA{\textit{Dep. of Networked Systems and Services} \\
    \textit{Budapest University of Tech. and Eco.} \\
    Budapest, Hungary \\
    pejo@crysys.hu}
    \and
    \IEEEauthorblockN{Marcell Frank}
    \IEEEauthorblockA{\textit{Fac. of Electrical Eng. and Informatics} \\
    \textit{Budapest University of Tech. and Eco.} \\
    Budapest, Hungary \\
    marcell.frank@edu.bme.hu}
    \and
    \IEEEauthorblockN{Krisztian Varga \& Peter Veliczky}
    \IEEEauthorblockA{\textit{Faculty of Natural Sciences} \\
    \textit{Budapest University of Tech. and Eco.} \\
    Budapest, Hungary \\
    \{kvarga,vepe\}@math.bme.hu}
    \thanks{Balazs Pejo is also affiliated with EGroup, Budapest, Hungary and 
    Gergely Biczok is also affiliated with HUN-REN-BME Information Systems Research Group, HUN-REN Hungarian Research Network, Office for Supported Research Groups, Budapest, Hungary}
}
\maketitle

\begin{abstract}
    This paper investigates the fragility of contribution evaluation in federated learning, a critical mechanism for ensuring fairness and incentivizing participation. We argue that contribution scores are susceptible to significant distortions from two fundamental perspectives: architectural sensitivity and intentional manipulation. First, we explore how different model aggregation methods impact these scores. While most research assumes a basic averaging approach, we demonstrate that advanced techniques, including those designed to handle unreliable or diverse clients, can unintentionally yet significantly alter the final scores. Second, we examine the threat posed by poisoning attacks, where malicious participants strategically manipulate their model updates to either inflate their own contribution scores or reduce others'. Through extensive experiments across diverse datasets and model architectures, implemented within the Flower framework, we rigorously show that both the choice of aggregation method and the presence of attackers can substantially skew contribution scores, highlighting the need for more robust contribution evaluation schemes.
\end{abstract}


\section{Introduction}
\label{sec:intro}

As digital technologies become integral to our daily routines, vast amounts of user data are collected, fueling the rise of Machine Learning (ML)~\cite{zhou2021machine}. However, relying on isolated, centralized datasets for training can lead to biased or suboptimal outcomes. Federated Learning (FL)~\cite{mcmahan2017communication} has emerged as a promising alternative, enabling multiple organizations or users to collaboratively train a model without sharing their raw data. This approach not only improves model quality by leveraging diverse data sources but also addresses critical privacy concerns by keeping data local.

Despite its advantages, FL is not a silver bullet and introduces significant challenges. One major issue is handling data heterogeneity across participants, which can reduce the effectiveness of standard aggregation techniques like Federated Averaging (FedAvg)~\cite{mcmahan2017communication}. To address this, advanced methods like FedProx~\cite{li2020federated} and FedNova~\cite{wang2020tackling} have been developed. Another critical issue is robustness against adversarial behavior. Malicious or ``Byzantine'' clients may send misleading updates to poison~\cite{tolpegin2020data} the model, hide a backdoor~\cite{bagdasaryan2020backdoor}, or disrupt the training process, compromising the model's integrity and reliability~\cite{shi2022challenges}. To counter these threats, Byzantine Fault Tolerant (BFT) robust aggregation techniques such as Krum~\cite{blanchard2017machine} and Zeno~\cite{xie2019Zeno} have been proposed~\cite{fung2018mitigating,guerraoui2018hidden}; though none offer a complete solution, especially when data is heterogeneously distributed across clients.

A third and increasingly important challenge is how to fairly evaluate the contributions of each participant. When the shared model has commercial value, the question of how to distribute the surplus becomes crucial. Without an effective reward scheme, the entire collaborative effort could collapse. Adapted from cooperative game theory, the Shapley Value (SV)~\cite{shapley1951notes} is often cited as the ideal solution for its fairness properties~\cite{winter2002shapley,rozemberczki2022shapley}, but its exponential computational cost makes it impractical. Consequently, various approximations and simpler methods have been developed to make Contribution Evaluation (CE) feasible~\cite{huang2020exploratory,jiang2023opendataval,liu2022gtg,wu2021fast}. A notable real-world example is the MELLODDY project for drug discovery~\cite{heyndrickx2023melloddy,oldenhof2023industry}, where competing pharmaceutical companies engage in ``co-opetition'' {zineldin2004co}, making fair and reliable CE essential for the collaboration's success.

This brings us to the core focus of this paper: the fragility of CE schemes. The reliability of these schemes is threatened from two distinct but related angles. First, there is an architectural sensitivity: CE methods are typically designed and tested with the standard FedAvg aggregator. However, as noted, real-world FL systems often require advanced aggregators to ensure robustness or handle data heterogeneity. This raises a fundamental question: how do these architectural improvements unintentionally distort contribution scores?

Second, the co-opetition dynamic in collaborative environments creates novel incentives for intentional manipulation. The objective of a malicious actor may not be to corrupt the global model, but rather to poison the contribution scores to unfairly gain a larger reward or make competitors lose standing. This motivates a new class of stealthy attacks designed to manipulate scores while leaving the model's performance largely intact. In this work, we investigate both phenomena, providing a comprehensive analysis of the fragility of contribution evaluation in modern federated learning systems.

\subsection*{Contribution}

In this work, we study the fragility of CE schemes in cross-silo FL from two distinct angles. Our key contributions are as follows:

\begin{itemize}
    \item \textbf{Architectural Sensitivity:} We identify a critical gap in the literature, where CE methods are almost exclusively tested with the standard FedAvg aggregator. 
    \begin{itemize}
        \item We conduct the first systematic analysis on how advanced aggregation methods (handling data heterogeneity and providing Byzantine Fault Tolerance) affect contribution scores.
        \item Our findings reveal that advanced aggregation techniques,  unintentionally but substantially, distort contribution scores across several state-of-the-art CE schemes. 
    \end{itemize}
    \item \textbf{Intentional Manipulation:} We design and implement two novel model poisoning attacks aimed specifically at manipulating contribution scores. 
    \begin{itemize}
        \item The \emph{Self Improvement} attack aims to increase the contribution score of the attacker, while \emph{Targeted Decrease} aims to reduce the score of a targeted client.
        \item Our results show that both attacks are effective in manipulating the contribution score of the selected client across various scenarios.
    \end{itemize}
    \item \textbf{Rigorous and Reproducible Evaluation:} We provide a comprehensive empirical validation of both the architectural and adversarial phenomena across multiple datasets, models, and contribution evaluation schemes. We have integrated all aggregation methods and attacks into the popular Flower FL framework~\cite{beutel2020flower}; our implementation is open-source to ensure reproducibility and encourage further research.
\end{itemize}

\subsection*{Organization}

The remainder of this paper is organized as follows. 
Section~\ref{sec:rw} reviews the related work.
In Section~\ref{sec:bg}, we recap the preliminaries on model aggregation and contribution scoring in FL.  
In Section~\ref{sec:mod}, we introduce our two novel contribution score poisoning attacks. 
Section~\ref{sec:exp} presents our experimental setup.
In Section~\ref{sec:res}, we i) evaluate the impact of different aggregation methods on contribution scores and ii) demonstrate the effectiveness of our adversarial attacks. 
Finally, we discuss the broader implications of our findings and conclude the paper in Section~\ref{sec:con}.
\section{Related Work}
\label{sec:rw}

Within this section, we situate our work within the existing literature by reviewing research at the intersection of contribution evaluation and Trustworthy ML. 

CE and Trustworthy ML are core pillars of FL, each with a vast scientific literature. However, they are mostly studied in isolation from each other. Here, we focus on their intersections and the gaps that motivate our work.

\subsection{Contribution Evaluation}

CE encompasses three distinct disciplines~\cite{rozemberczki2022shapley}: \textit{explainability}~\cite{gunning2019xai}, which scores data features; \textit{data evaluation}~\cite{wang1996beyond}, which evaluates individual data points; and \textit{contribution score computation}~\cite{wang2019measure}, which scores the client datasets. Prior work has explored the intersection of CE with privacy~\cite{luo2022feature,ezzeddine2024privacy,watson2022differentially,wang2023threshold,zheng2022secure,pejo2023quality} and fairness~\cite{huang2021shapley,pandl2022reward}. A handful of works have also considered its relation to robustness; some studied the robustness of the SV scores themselves~\cite{wang2023data,goldwasser2024stabilizing}, while others used CE techniques to create robust aggregation mechanisms~\cite{otmani2024fedsv}. However, this line of work does not address the architectural sensitivity of CE scores; the question of how the choice of an advanced aggregation method unintentionally impacts score distributions remains largely unexplored. Our paper focuses on client-level contribution score computation and its fragility to both architectural choices and direct manipulation.

\subsection{Trustworthy ML}

Traditionally, poisoning attacks have focused on reducing model performance, either targeting specific classes or degrading performance overall~\cite{ibitoye2019threat,goldblum2022dataset,xia2023poisoning}. Besides, poisoning has been leveraged to manipulate other model properties, such as exacerbating privacy leakage~\cite{lecuyer2019certified,ma2019data,nasr2018machine,song2019privacy,lyu2022privacy}, manipulating explanations~\cite{schupbach2018robustness,hancox2020robustness,hsieh2020evaluations,artelt2021evaluating}, or increasing fairness disparities~\cite{nanda2021fairness,xu2021robust,benz2021robustness,shi2023towards}. While these attacks do not directly aim to reduce accuracy, they often impact model performance indirectly. Our work introduces attacks where the primary goal is not to harm the model but to manipulate the metadata of the training process, namely, the contribution scores themselves.

\subsection{Adversarial Contribution Evaluation}

The adversarial manipulation of contribution scores is a nascent field. We are aware of only three works that study this interaction without focusing on model performance degradation~\cite{arnaiztowards,anada2025measuring,xu2024ace}. Arnaiz-Rodriguez et al.~\cite{arnaiztowards} used CE to improve model fairness, Anada et al.~\cite{anada2025measuring} focused on decentralized FL with self-reported scores, while Xu et al.~\cite{xu2024ace} presented the ACE attack to increase a client's contribution score without training on good-quality data. Our work follows this latter direction. 

The ACE attack~\cite{xu2024ace} was the first designed to inflate a contribution score by predicting the aggregated model in the next round using first and second-order optimization. While novel, the authors only studied a limited scenario: increasing the attacker's own score using a rudimentary CE scheme based on cosine distance. In contrast, we develop algorithms for multiple adversarial goals (both increasing the attacker's own score and decreasing the target's) and evaluate them against principled, Shapley-inspired CE schemes based on marginal difference. 

\subsection{Additional Literature Review}

Given that ML is a fast-moving field and our topic lies at the intersection of multiple sub-fields, we reviewed recent surveys~\cite{zhang2023survey,cui2024survey} and conducted a broader literature search via Google Scholar to confirm that no prior work explicitly addresses the manipulation of FL contribution scores.
The utilized queries were ``\textit{``Contribution'' and ``Poison'' and ``Federated Learning''}'' and ``\textit{``Contribution Evaluation'' and ``Aggregation'' and ``Federated Learning''}''. Regarding CE poisoning, our review did not find papers not referenced previously. Regarding aggregation methods, we found several studies (e.g.,~\cite{zeng2021fedcav,du2025hfedcwa}) proposing to use contribution scores for designing more robust/fair aggregation mechanisms; however, the reverse relationship (i.e., how the choice of the aggregation mechanism affects the resulting contribution scores) seems to remain unexplored. 
\section{Background}
\label{sec:bg}

This section recaps the preliminaries that form the foundation of our study. We first formalize the FL process and its main challenges. We then provide a detailed overview of the advanced aggregation methods and contribution evaluation schemes central to our analysis. Table~\ref{tab:not} summarizes the notations used throughout the paper.

\subsection{Federated Learning}

Let us first recall the formalization of the general FL problem. We have $K$ clients, each with their own local dataset $\mathcal{D}_k$ of size $n_k$, for $k \in [K]$. Each client $k$ seeks to minimize its local objective, the empirical loss over its dataset $\mathcal{D}_k$ as described on the left of Equation~\ref{eq:FL}, where $\ell(\omega, x)$ is the loss function for data point $x$ with model parameters $\omega$. The training proceeds for $T$ rounds, where the server periodically aggregates the local models with the default FedAvg~\cite{mcmahan2017communication} mechanism, which in round $t$ takes the weighted average of the client model updates according to their corresponding dataset sizes, as described on the right side of Equation~\ref{eq:FL}. Notations without a subscript correspond to the server, i.e., $\omega^t$ is the aggregated model after round $t$, $\mathcal{D}$ is the validation dataset, etc. 

{\small
\begin{equation}
    \label{eq:FL}
    f_k(w_k^t) = \frac{1}{n_k} \sum_{x \in \mathcal{D}_k} \ell(w_k^t, x)
    \hspace{1cm}
    \omega^t = \sum_{k=1}^{K} \frac{n_k}{n} \omega^t_k
\end{equation}}

While FL offers significant privacy and data diversity benefits, its distributed nature introduces critical challenges to performance and reliability. Notably, client heterogeneity, where differences in local data and computational systems hinder model convergence, is a primary concern. This decentralization also creates opportunities for malicious behavior from Byzantine attackers, who can disrupt training through data poisoning (manipulating local datasets) or model poisoning (altering model updates before transmission)~\cite{shi2022challenges}.

\subsection{Heterogeneity-Aware Aggregation}

To address client diversity, specialized aggregation techniques have been developed; within this work, we consider two well-established methods.

\vspace{0.1cm}\subsubsection{FedProx~\cite{li2020federated}}

FedProx modifies the local training objective $f_k(\omega)$ by introducing a proximal term with parameter $\mu$, as seen on the left side of Equation~\ref{eq:hetero}. This term prevents local models from deviating far from the global model, thereby stabilizing training when clients operate on non-IID data.

\vspace{0.1cm}\subsubsection{FedNova~\cite{wang2020tackling}}

FedNova aims to deal with uneven computational resources by normalizing client updates. This is done by adjusting the updates based on their local workload, such as the number of locally performed training rounds $\tau_k$. The global model update in round $t$ is computed by aggregating the normalized local updates, as shown on the right side of Equation~\ref{eq:hetero}. Here $\Delta\omega_k^t$ denotes the update vector of client $k$ at round $t$.

{\footnotesize
\begin{equation}
    \label{eq:hetero}
    \min_{\omega_k^t} \left( f_k(\omega_k^t) + \frac{\mu}{2} \|\omega_k^t - \omega^{t-1}\|^2 \right)
    \hspace{0.5cm}
    \omega^{t} = \omega^{t-1} - \sum_{k=1}^{K} \frac{\tau_k n_k}{\tau n}\Delta\omega_k^t
\end{equation}}

{\small
\begin{table}[t]
    \centering 
    \caption{Notations used in the paper.}
    \label{tab:not}
    \begin{tabular}{|c|m{6.2cm}|}
    \hline
    \textbf{Symbol} & \textbf{Description} \\ \hline\hline
    $K,T$ & Number of clients \& training rounds. \\ \hline
    $\mathcal{D}_k,f_k$ & Dataset \& objective function of client $k$. \\ \hline
    $n_k,n$ & Size of $\mathcal{D}_k$ \& the total number of samples. \\ \hline
    $\omega^t_k,\omega^t$ & 
    Local and global model weights of client $k$ \linebreak after round $t$ (when no CE poisoning is present). \\ \hline
    $\omega^t_a,\hat{\omega}^t$ & 
    Poisoned model weights of attacker $a$ and \linebreak the affected global weights after round $t$. \\ \hline
    $\eta, \lambda, \nu$ & Client learning rate, learning rate decay, and (local) momentum. \\ \hline\hline
    $\mu$ & Proximal term parameter for FedProx. \\ \hline
    $\tau_k, \tau$ & Num. of local training rounds of client $k$ \& total num.\\ \hline\hline
    $\kappa$ & Number of tolerated Byzantine clients. \\ \hline
    $\rho$ & Regularization parameter in Zeno aggregation. \\ \hline\hline
    $\phi_v^t(k)$ & Contribution score of client $k$ in round $t$ based on evaluation function $v$. $\phi\in\{GTG,LOO,ADP\}$. \\ \hline\hline
    $\gamma$ & Weight for regularization in Targeted Decrease \\ \hline
    \end{tabular}
\end{table}}

\subsection{Byzantine Fault Tolerant Aggregation}

These defense mechanisms modify the aggregation process to mitigate the impact of malicious updates. In this paper, we consider two well-known mechanisms.

\vspace{0.1cm}\subsubsection{Krum~\cite{blanchard2017machine}}

Krum is one of the first robust aggregation mechanisms. It assigns a score to each client based on the sum of squared Euclidean distances to its $K-\kappa-2$ nearest neighbors, where $\kappa$ is the number of expected attackers. Denoting with $k^\prime$ the clients whose update $\omega^t_{k^\prime}$ belongs to the set of the nearest vectors $\omega^t_k$, Krum selects the single client update with the smallest such score as the global update, as presented on the left side of Equation~\ref{eq:bft}.

\vspace{0.1cm}\subsubsection{Zeno~\cite{xie2019Zeno}}

Zeno uses a stochastic zero-order oracle to evaluate client updates by measuring the loss reduction on a trusted dataset $\mathcal{D}$. It scores each client based on this estimated descent, penalized by a regularization term $\rho$, as formulated in Equation~\ref{eq:zeno}. Zeno then aggregates the updates from the top $K-\kappa$ clients with the highest scores as shown in the right side of Equation~\ref{eq:bft}. Here $n_{K^\prime}$ is the sum of $n_k$ for $k$ in $K^\prime$.

{\small
\begin{equation}
    \label{eq:bft}
    \omega^t = \omega^t_{\arg\min_k\left(\sum_{k^\prime} \|\omega_k^t - \omega_{k^\prime}^t\|_2^2\right)}
    \hspace{0.5cm}
    \omega^t = \frac{1}{|K^\prime|}\sum_{k\in K^\prime} \frac{n_k}{n_{K^\prime}} \omega^t_k
\end{equation}}

{\footnotesize
\begin{equation}
    \label{eq:zeno}
    K^\prime=\underset{K-\kappa}{\arg\text{top}}\phantom{}_k\left(\frac1{|\mathcal{D}|}\sum_{x\in \mathcal{D}}\left(\ell(\omega^{t-1}, x)-\ell(\omega^t_k, x)\right)-\rho\|\omega^t_k\|^2_2\right)
\end{equation}}

\subsection{Contribution Evaluation}

Unlike misbehavior detection, which categorizes clients as either honest or malicious, CE assigns nuanced, continuous scores that quantify each client's impact on the global model. Here we consider three such schemes: one distance-based and two based on marginal differences. 

\vspace{0.1cm}\subsubsection{The Shapley Value (SV)~\cite{shapley1951notes}}

The Shapley Value originates from cooperative game theory and is widely regarded as a principled method for evaluating individual contributions in collaborative settings. Its strength lies in provably being the only reward allocation scheme that satisfies four key fairness axioms simultaneously (efficiency, symmetry, null player, and additivity). It computes a score for a client based on its average marginal contribution to all possible subsets (coalitions) of other clients. Formally, for a set of $K$ clients and a utility function $v$, the SV for client $k$ is presented in Equation~\ref{eq:shap_original}.

{\small
\begin{equation}
    \label{eq:shap_original}
    SV_v(k) = \frac{1}{K} \sum_{S \subseteq [K] \setminus \{k\}} \frac{v(S \cup \{k\}) - v(S)}{\binom{K-1}{|S|}}
\end{equation}}

For instance, in the FL setup, the evaluation function $v$ can be the accuracy or the loss of the trained models. Calculating the exact SV is expensive, as it requires an exponential number of trained models. This computational burden was significantly reduced by the \emph{Contribution Index}~\cite{song2019profit}, which shifted the overhead from training to inference by evaluating combinations of model updates in each round. These round-wise scores are then aggregated into the final score. For notational convenience, we denote any round-specific value with a superscript $t$, and omit the superscript for the final aggregated value.

\vspace{0.1cm}\subsubsection{GTG Shapley~\cite{liu2022gtg}}

Guided Truncated Gradient Shapley (GTG) further reduces the required number of model evaluations by using an alternative, permutation-based form for the SV, as shown in Equation~\ref{eq:shap_permute}.

{\small
\begin{equation}
    \label{eq:shap_permute}
    SV_v(k) = \frac{1}{K!} \sum_{\pi \in \Pi(K)} \left( v(\pi_{-k} \cup \{k\}) - v(\pi_{-k}) \right) 
\end{equation}}

Here, instead of subsets, a coalition is formed one client at a time, and the average marginal contribution is taken over the different permutations $\Pi(K)$. The notation $\pi_{-k}$ is the set of clients in permutation $\pi$ which precede client $k$. The computational reduction of GTG is three-fold:

\begin{enumerate}
    \item It disregards entire rounds if the overall utility gain is small (e.g., if the evaluation function is loss, round $t$ is omitted when $v(\pi^{t-1})-v(\pi^t)<\epsilon_0$).
    \item It considers only a subset $\Pi^\prime(K)$ of all possible $\Pi(K)$ permutations, where $|\Pi^\prime(K)|/|\Pi(K)|<\epsilon_1$.
    \item It only considers the prefix of each selected permutation, where $v(\pi_{-k}\cup \{k\}) - v(\pi_{-k}) > \epsilon_2$.
\end{enumerate}

Moreover, the selection of $\Pi^\prime(K)$ is governed by a guided sampling method which ensures that each client is uniformly present in the prefixes of the selected permutations. 

\vspace{0.1cm}\subsubsection{Leave One Out (LOO)~\cite{evgeniou2004leave}}

We also consider a linear approximation of the SV. While GTG is a state-of-the-art approximation, LOO is more lightweight and widespread. The contribution is computed as the difference between the utility of the model aggregated with all clients and the utility of the model aggregated with all clients except for client $k$, as shown in Equation~\ref{eq:loo}. 

{\small
\begin{equation}
    \label{eq:loo}
    LOO_v(k) = v([K]) - v([K]\setminus\{k\})
\end{equation}}

\vspace{0.1cm}\subsubsection{Adaptive Weighting (ADP)~\cite{wu2021fast}}

In contrast to Shapley-based scores, ADP is an efficient, non-marginal-difference-based method that utilizes distances, more specifically, the cosine similarity between local model updates and the aggregated global update. Furthermore, it smooths these scores over time to reflect the historical contribution of each client, giving a higher score to clients whose updates consistently align with the global objective. The exact formula is in Equation~\ref{eq:fedadp}, where $\theta^0_k=0$.

{\small
\begin{equation}
    \label{eq:fedadp}
    \theta_k^{t+1} = \frac{t\cdot\theta_k^t + 1-\cos(\angle(\omega^t_k,\omega^t))}{t+1}
    \hspace{0.4cm}
    ADP^t(k) = \frac{1 - e^\frac{-1}{\theta_k^t}}{1 + e^\frac{-1}{\theta_k^t}}
\end{equation}}
\section{Threat models, attacks, and system formalization}
\label{sec:mod}

In this section, we present i) our envisioned scenarios along with the corresponding threat models, ii) show how an adversarial client can craft its model update to manipulate the contribution scores (increasing its own or decreasing someone else's) computed by the server, and iii) formalize our system model. The notations used are summarized in Table~\ref{tab:not}. 

\subsection{Threat Model and Attacker Objectives}

\vspace{0.1cm}\subsubsection{Architectural Sensitivity}

In our first scenario, we consider an FL setting with an \emph{honest-but-curious} server: it executes the training protocol exactly as specified but may attempt to learn additional information from aggregated values. In the meantime, most clients are considered \emph{honest}, while some may be \emph{malicious}; therefore, robust aggregation is required. Our objective is to analyze how the server's choice of aggregation algorithm systematically alters the distribution of the final scores, even when only benign clients are present.

\vspace{0.1cm}\subsubsection{Intentional Manipulation}

In our second scenario, we assume that the majority of clients are \emph{honest}, but a subset of clients may be \emph{rational}: their actions are guided by an underlying utility function which they aim to maximize. We identify three main factors that can be balanced within this function: the overall model quality, the adversary’s own contribution score, and the contribution scores of other clients.

Depending on how these weights are set, multiple adversarial strategies can emerge. In this paper, we focus on two representative cases where the selfish client can submit a carefully crafted, poisoned model update $\omega_a^t$ in place of their honestly computed update using $\mathcal{D}_a$:

\begin{itemize}
    \item \textbf{Self Improvement:} In round $t$, the adversary $a$ aims to craft an update $\omega_a^t$ to maximize their own score, $\phi_v^t(a)$. 
    \item \textbf{Targeted Decrease:} In round $t$, the adversary $a$ aims to craft an update $\omega_a^t$ to minimize the score of a targeted benign client $k$, $\phi_v^t(k)$.
\end{itemize}

We assume the adversary has access to the server's test set. This is a common assumption, as the clients usually know what benchmark their assigned contribution score is based on. Moreover, the attacker also has access to the updates of benign clients prior to the final aggregation and contribution evaluation. Although this represents a hypothetical worst-case scenario, it can be realized in several ways. For instance, the server can be \emph{malicious}, i.e., the rational client could host the server where only the CE computations are verifiable, as that is more lightweight than the actual training. Thus, the adversary can craft any gradient, but cannot tamper with the CE computation. Alternatively, the server could be \emph{honest}, but the rational client could collude with the network provider and obtain this information via eavesdropping.

\subsection{System Formalization}

Following the notations used in Section~\ref{sec:bg}, we consider a standard cross-silo FL setting where the local model updates $\omega_k^t$ are aggregated by the server to produce in each round the global model, $\omega^t$. Following the aggregation step, CE scheme $\phi$ is applied. This function takes the set of local updates $\{\omega_1^t, \dots, \omega_K^t\}$ as input and outputs a vector of contribution scores of the clients for that round. In general, the final contribution scores of the clients are the aggregate (e.g., average) of these round-wise scores. Hence, in our Architectural Sensitivity study, we focus on the final scores, as we are interested in the overall effect of the aggregation over all rounds. On the other hand, for Intentional Manipulation, our attacks focus on a single round of FL: it is enough to demonstrate the effectiveness of score manipulation within a single round. Clearly, the attacker can apply these techniques iteratively and achieve an even larger impact.  

An overview of the FL process is shown in Figure~\ref{fig:overview}: within a round, the server broadcasts the previous aggregated model, which the honest clients train further according to the prescribed protocol, while the dishonest clients carry out an attack. These model updates are then sent back to the server which, besides aggregating them with the appropriate mechanism, invokes the desired CE mechanism. The general process for a single training round is also encapsulated in Alg.~\ref{alg:poison}. 

{\small
\begin{figure}[t]
    \centering
    \caption{Our FL scenarios: Architectural Sensitivity (marked with a 5-star) uses all aggregation techniques and GTG and ADP as Contribution Evaluation, while Intentional Manipulation (marked with a 6-star) relies on FedAvg and considers GTG and LOO.} 
    \label{fig:overview}    \includegraphics[width=0.45\textwidth]{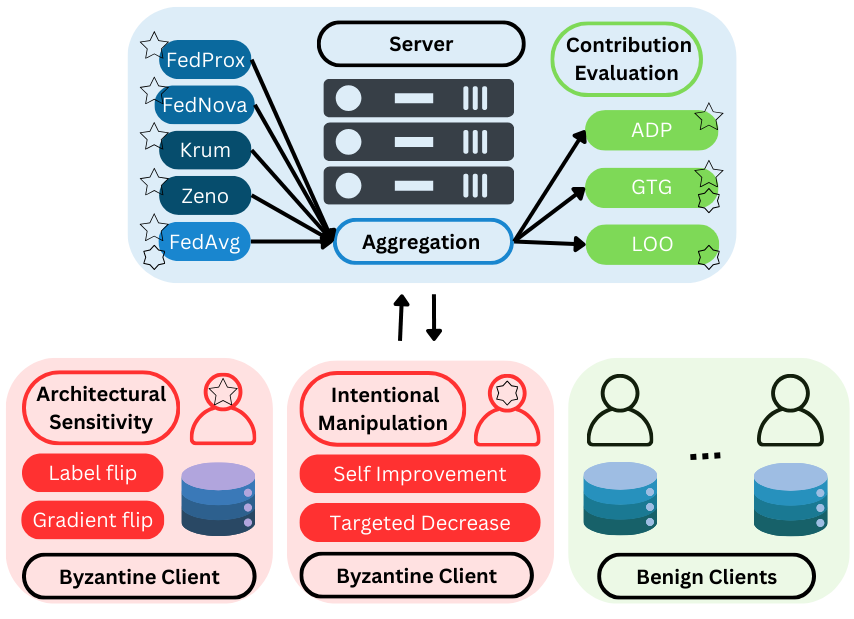}
\end{figure}}

{\small
\begin{algorithm}[!t]
    \caption{A single training round of the FL process}
    \label{alg:poison}
    \textbf{Input}: previous global model $\omega^{t-1}$; aggregation mechanism $agg$; training method $Train$; attack technique $Attack$; CE scheme $\phi$ \\
    \textbf{Output}: new global model $\omega^t$; contribution scores $\{\phi^t(1),\dots,\phi^t(K)\}$
    \begin{algorithmic}[1]
    \State Server sends $\omega^{t-1}$ to clients.
    \For{each client $k \in [K] \setminus \{a\}$}
        \If{client is benign}
            \State Client $k$ computes its honest update:
            \Statex\quad\quad\quad $\omega_k^t \leftarrow Train(\omega^{t-1}, \mathcal{D}_k)$.
            % TODO: Fix indentation
        \Else
            \State Client $k$ executes Byzantine attack: 
            \Statex\quad\quad\quad $\omega_k^t \leftarrow Attack(\omega^{t-1})$.
            % NOTE: Doesn't Attack also depend on \mathcal{D}_k?
        \EndIf
        \State Client $k$ sends $\omega_k^t$ to the server.
    \EndFor
    \If{Intentional Manipulation}
        \State Adversary $a$ obtains client updates $\{\omega_k^t\}_{k \neq a}$.
        \State Adversary $a$ computes a poisoned update $\omega_a^t$:
        \If{Self Improvement}
            \State $\omega_a^t\leftarrow$ Run optimization Equation~\ref{eq:ce_si}
        \EndIf
        \If{Targeted Decrease}
            \State $\omega_a^t\leftarrow$ Run optimization Equation~\ref{eq:ce_td}
        \EndIf
        \State Adversary $a$ sends $\omega_a^t$ to the server.
        \EndIf
    \State Server aggregates all received updates: 
    \Statex $\omega^t \leftarrow Agg(\{\omega_1^t, \dots, \omega_K^t\})$.
    \State Server computes CE scores for all clients: 
    \Statex $\left(\{\phi^t(1), \dots, \phi^t(K)\}\right) \leftarrow \phi\left(\{\omega_1^t, \dots, \omega_K^t\}\right)$.
    \end{algorithmic}
\end{algorithm}} 

\vspace{0.1cm}\subsubsection{Self Improvement}

The adversary's goal is to inflate their own contribution score. To achieve this, the adversary reverse-engineers the evaluation metric and crafts an update by solving the unconstrained optimization problem shown on the left side of Equation~\ref{eq:ce_si}, which aims to directly maximize their own score. This problem can be solved using standard gradient ascent, but its success hinges on a strong assumption: the adversary must have knowledge of the server's validation dataset, $\mathcal{D}$, and the specific CE function, $\phi$, being used.

{\small
\begin{equation}
    \label{eq:ce_si}
       \max_{\omega_a^t} \phi^t(a) 
        \hspace{0.5cm}\Rightarrow\hspace{0.5cm}
        \min_{\omega_a^t}f_a(w_a^t)=\min_{\omega_a^t} \frac1{n_a}\sum_{x_i\in\mathcal{D}_a}\mathcal{L}(\omega_a^t;x_i)
\end{equation}}

Since higher scores are awarded for updates that improve the global model, this optimization has the positive side-effect of also minimizing the global loss. Putting it differently, the selfish act of maximizing a reward indirectly results in a high-quality model. This phenomenon, where an attack on the scoring mechanism yields a beneficial model update, was also observed in the context of the ACE attack~\cite{xu2024ace} for a less principled CE scheme without a game-theoretic grounding. The above blurs the line between a malicious attack and a legitimate, albeit unconventional, optimization strategy. In fact, such a maneuver should be perceived as an attack on the \emph{fairness} of the reward system, but not on the model's performance itself. 

\vspace{0.1cm}\subsubsection{Targeted Decrease}

In this direct attack against a competitor, the adversary's goal is to minimize the contribution score of a specific target client, $\hat{k}$. Unlike the inflation attack, this objective is fundamentally at odds with improving the global model. Minimizing a benign client's score, which reflects their positive impact, inherently risks degrading the model's overall performance. To overcome this, the adversary must solve the constrained optimization problem shown in Equation~\ref{eq:ce_td}, which also includes a regularization term. The objective is to find a poisoned update $\omega_a^t$ that minimizes the target's score, $\phi^t(\hat{k})$, while adhering to a strict performance constraint. This constraint ensures that the collateral damage to the global model's loss, compared to a baseline model aggregated without the attacker, is bounded by a small threshold $\varepsilon$. Here, $\omega^t$ is the final global model including the attacker's poisoned update, while $\hat{\omega}^t$ represents the model aggregated using only the benign clients' updates.

{\small
\begin{equation}
    \label{eq:ce_td}
        \min_{\omega_a^t} \left[\phi^t\left(\hat{k}\right)+\gamma\cdot\|\omega_a^t\|_2^2\right]
        \hspace{0.5cm}\text{s.t.}\hspace{0.5cm}
        \|f(\omega^t) - f(\hat{\omega^t})\|_2^2 < \varepsilon
\end{equation}}
\section{Experimental Setup}
\label{sec:exp}

This section describes our experimental setup. First, we outline the common aspects across all experiments; then, we detail the specific configurations for our two distinct scenarios. A high-level summary of the two scenarios is provided in Table~\ref{tab:exp_summary}.

\subsection{Common Aspects}

All experiments are implemented using the Flower framework~\cite{beutel2020flower} with PyTorch. We simulate a setting with only $K=5$ clients for two reasons. First, most CE scores lose their descriptive power with a large(r) number of participants, as the majority of the contribution values tend to converge to zero. Second, in real-world cross-silo FL deployments (e.g.,~\cite{heyndrickx2023melloddy}), participation is typically limited to only a handful of parties, as legal and technical constraints hinder collaboration at larger scales. We evaluate on both IID and non-IID data distributions, where the non-IID setting is created using a Dirichlet distribution over the class labels (and when applicable, also over the computational resources). All experiments are repeated 10 times to ensure statistically relevant results. Our code is open-source under the MIT License: \url{https://anonymous.4open.science/r/fragility_of_CE_combined-23F1}.

%\footnote{\url{https://github.com/Varga-Krisz/BYZvsCE}}\footnote{\url{https://github.com/m9framar/conteval-pois-Flower}}

\subsection{Architectural Sensitivity}

We investigate how different aggregation rules affect CE scores under three distinct conditions. First, to test heterogeneity-aware aggregators (FedProx, FedNova), we use various non-IID data distributions created with a Dirichlet process ($\alpha=1.0$ and $\alpha=0.1$). Second, to test Byzantine Fault Tolerant aggregators (Krum, Zeno), we introduce a simple baseline adversary performing label- or gradient-flipping on an IID distribution. Finally, to assess the impact of gradually degrading data quality, we create a linear label noise scenario. In this setup, for each client $k \in \{1, \dots, K\}$, we corrupt its local dataset by randomly re-assigning the label of each sample with a probability $p_k$, where the noise level increases linearly with the client's index: $p_k=(k-1)/(K-1).$ This creates a controlled environment where Client 1 has unchanged labels while Client $K$ has completely random labels. Across all scenarios, client contributions are evaluated using both the marginal difference-based GTG and the distance-based ADP. Since these scores may lie on different scales (and can even be negative), we apply minimum-offset correction followed by normalization. The resulting values lie in $[0,1]$ and sum to $1$.

To ensure a fair comparison, the client-side optimizer (SGD with momentum and decay) is independently fine-tuned for each aggregation method based on its intended operational scenario. We perform a grid search over the learning rate ($\eta \in \{0.1, 0.01, 0.001\}$), LR decay ($\lambda \in \{0.97, 0.99, 1.0\}$), and momentum ($\nu \in \{0.0, 0.5, 0.9\}$). Specifically, FedAvg is optimized on IID data, FedProx and FedNova on highly non-IID data ($\alpha=0.1$), and Krum and Zeno on IID data with an assumption of two Byzantine clients. Key aggregation-specific parameters are set to standard values from their respective literature: $\mu=1.0$ for FedProx and $\rho=100$ for Zeno. For GTG, the evaluation function is the model's loss with approximation thresholds set to $\epsilon_0=0.0002, \epsilon_1=0.75$, and $\epsilon_2=0.0001$.

{\small
\begin{table}[t]
    \centering
    \caption{Summary of Experimental Setups.}
    \label{tab:exp_summary}
        \begin{tabular}{|c||m{3cm}|m{3cm}|}
        \hline
        & \textbf{Architectural \linebreak Sensitivity} & \textbf{Intentional \linebreak Manipulation} \\ \hline \hline
        \textbf{Data} & CIFAR-10 & ADULT \linebreak Fashion-MNIST \\ \hline
        % NOTE: ADULT (+ MLP) / CIFAR-100 for AS?
        \textbf{Dist.} & IID, IID + noise \hfill \hfill \linebreak non-IID high \& low imb. & IID \linebreak non-IID low imbalance \\ \hline
        \textbf{Arch.} & 2-layer CNN & 3-layer MLP \hfill \hfill \linebreak 2-layer CNN \\ \hline
        \textbf{Agg.} & Fed -Avg, -Prox, -Nova \linebreak Krum, Zeno & FedAvg \\ \hline
        \textbf{Adv.} & Only benign \hfill \hfill \linebreak Label \& Gradient flip & Self Improvement \hfill \hfill \linebreak Targeted Decrease \\ \hline
        \textbf{CE} & GTG, ADP & GTG, LOO \\ \hline  % NOTE: LOO for AS?
    \end{tabular}
\end{table}} 

\subsection{Intentional Manipulation}

We demonstrate the feasibility of our contribution score poisoning attacks across two distinct tasks: tabular classification on the ADULT dataset with an MLP, and image classification on the Fashion-MNIST dataset with a CNN. The corresponding regularization terms and the threshold in Equation~\ref{eq:ce_td} are $0.001$ and $0.005$, respectively. We execute both attacks in IID and non-IID ($\alpha=1.0$) settings, focusing our analysis on a single, representative round of FL. Our attacks specifically target principled, marginal difference-based CE schemes (GTG and LOO), as the vulnerability of simpler, non-marginal methods has already been demonstrated by the ACE attack~\cite{xu2024ace}.

A key enabling aspect of this analysis is the assumption on the attacker's knowledge of the server's validation set. The \emph{Targeted Decrease} attack proved effective already with only a small fraction (1-3\%) of the validation set. The \emph{Self Improvement} attack, however, required the full validation set to succeed, limiting its application to the smaller ADULT dataset due to memory constraints. 
\section{Results}
\label{sec:res}

In this section, we present the empirical results of our two-pronged investigation into the fragility of contribution scores. We first analyze the Architectural Sensitivity of scores by evaluating how different aggregation methods impact the resulting score distributions. We then evaluate the effectiveness of our two proposed Intentional Manipulation attacks and discuss the practical implications.

\subsection{Architectural Sensitivity}

\vspace{0.1cm}\subsubsection{Heterogeneity-Aware Aggregation}

We first evaluated the effect of aggregators designed to handle non-IID data (FedProx and FedNova). By analyzing the scores produced under data imbalance (detailed in Table~\ref{tab:AS_noniid}), we observed a clear divergence in how the CE methods operate. The non-marginal ADP scores closely followed the clients' data ratio distributions, suggesting it primarily measures contribution quantity in this setting. In contrast, the marginal difference-based GTG scores were significantly more stable, with most scores falling within a tight range of $[0.18, 0.22]$ despite the significant data imbalance, indicating a focus on contribution quality.

The statistical analysis validates these observations. We employed the Anderson–Darling test to assess whether the score distributions under a given aggregation method are consistent with those obtained with FedAvg. The tests (Table~\ref{tab:AS_test}) show that ADP scores are statistically different from the FedAvg baseline in non-IID scenarios. This is especially true in the high imbalance setting: RMSE (Table~\ref{tab:AS_agg}) values also support this sizable difference. GTG scores, however, remain comparatively robust to the change in aggregation method, with significant differences only appearing for a few clients in the low-imbalance ($\alpha=1.0$) setting. This is a notoriously challenging scenario for contribution evaluation, as the data distribution is close to IID, where clients are expected to have similar scores.

\begin{table}[t]
    \centering
    \caption{CE scores with non-IID clients.}
    \label{tab:AS_noniid}
    \resizebox{.47\textwidth}{!}{%
    \begin{tabular}{ll||c|c|c|c|c}
	\toprule
        \multicolumn{2}{c||}{\textbf{High}} & \textbf{Cli.\,1} & \textbf{Cli.\,2} & \textbf{Cli.\,3} & \textbf{Cli.\,4} & \textbf{Cli.\,5} \\ 
        \multicolumn{2}{c||}{\textbf{(data ratios)}} & \textbf{18.5\%} & \textbf{51.1\%} & \textbf{14.0\%} & \textbf{6.1\%} & \textbf{10.3\%} \\
        \midrule
    
        \multirow{2}{*}{\textbf{FedAvg}}
        & \textbf{ADP} &.170 ± .000&.556 ± .001&.128 ± .000&.054 ± .000&.091 ± .000\\
        & \textbf{GTG} &.201 ± .022&.187 ± .030&.195 ± .019&.198 ± .028&.219 ± .028\\
        \midrule
        
        \multirow{2}{*}{\textbf{FedProx}}
        & \textbf{ADP} &.171 ± .001&.556 ± .001&.128 ± .001&.054 ± .000&.091 ± .000\\
        & \textbf{GTG} &.184 ± .018&.215 ± .03&.203 ± .024&.206 ± .021&.191 ± .018\\
        \midrule
        
        \multirow{2}{*}{\textbf{FedNova}}
        & \textbf{ADP} &.170 ± .000&.556 ± .000&.128 ± .000&.054 ± .000&.091 ± .000\\
        & \textbf{GTG} &.189 ± .017&.207 ± .022&.176 ± .025&.205 ± .029&.223 ± .027\\
        \midrule
        
        \midrule
        \multicolumn{2}{c||}{\textbf{Low}} & \textbf{Cli.\,1} & \textbf{Cli.\,2} & \textbf{Cli.\,3} & \textbf{Cli.\,4} & \textbf{Cli.\,5} \\
        \multicolumn{2}{c||}{\textbf{(data ratios)}} & \textbf{21.6\%} & \textbf{22.7\%} & \textbf{23.7\%} & \textbf{15.0\%} & \textbf{17.0\%} \\
        \midrule
        
        \multirow{2}{*}{\textbf{FedAvg}}
        & \textbf{ADP} &.217 ± .000&.227 ± .000&.245 ± .001&.142 ± .000&.169 ± .000\\
        & \textbf{GTG} &.196 ± .018&.192 ± .025&.197 ± .03&.208 ± .026&.206 ± .036\\
        \midrule
        
        \multirow{2}{*}{\textbf{FedProx}}
        & \textbf{GTG} &.212 ± .017&.191 ± .028&.216 ± .016&.192 ± .017&.188 ± .040\\
        & \textbf{ADP} &.217 ± .001&.228 ± .001&.244 ± .003&.142 ± .002&.169 ± .001\\
        \midrule
        
        \multirow{2}{*}{\textbf{FedNova}}
        & \textbf{ADP} &.217 ± .000&.227 ± .001&.243 ± .003&.144 ± .003&.169 ± .001\\
        & \textbf{GTG} &.199 ± .017&.208 ± .026&.207 ± .021&.196 ± .042&.191 ± .022\\
        \bottomrule
    \end{tabular}}

    \vspace{0.5cm}
    \caption{CE scores with label and gradient flip attackers.}
    \label{tab:AS_attack}
    \resizebox{0.45\textwidth}{!}{%
    \begin{tabular}{ll||c|c|c|c|c}
        \toprule
        \multicolumn{2}{c||}{\textbf{Label}} & \textbf{Cli. 1} & \textbf{Cli. 2}& \textbf{Cli. 3}& \textbf{Cli. 4}& \textbf{Att.} \\
        \midrule
        
        \multirow{2}{*}{\textbf{FedAvg}}
        & \textbf{ADP} &.203 ± .000&.203 ± .000&.203 ± .000&.203 ± .000&.188 ± .001\\
        & \textbf{GTG} &.198 ± .060&.198 ± .038&.191 ± .039&.222 ± .047&.192 ± .041\\
        \midrule
        
        \multirow{2}{*}{\textbf{Krum}}
        & \textbf{ADP} &.200 ± .005&.208 ± .006&.209 ± .007&.211 ± .007&.173 ± .001\\
        & \textbf{GTG} &.194 ± .036&.201 ± .039&.207 ± .04&.200 ± .033&.198 ± .038\\
        \midrule
        
        \multirow{2}{*}{\textbf{Zeno}}
        & \textbf{ADP} &.207 ± .002&.208 ± .002&.206 ± .002&.208 ± .002&.172 ± .001\\
        & \textbf{GTG} &.188 ± .051&.186 ± .033&.207 ± .043&.209 ± .050&.210 ± .039\\
        \midrule
        
        \midrule
        \multicolumn{2}{c||}{\textbf{Grad.}} & \textbf{Cli. 1} & \textbf{Cli. 2}& \textbf{Cli. 3}& \textbf{Cli. 4}& \textbf{Att.} \\
        \midrule
        
        \multirow{2}{*}{\textbf{FedAvg}}
        & \textbf{ADP} &.214 ± .001&.214 ± .001&.214 ± .001&.214 ± .001&.145 ± .002\\
        & \textbf{GTG} &.201 ± .023&.207 ± .023&.206 ± .030&.202 ± .037&.184 ± .042\\
        \midrule
        
        \multirow{2}{*}{\textbf{Krum}}
        & \textbf{ADP} &.210 ± .007&.213 ± .006&.216 ± .007&.219 ± .003&.141 ± .002\\
        & \textbf{GTG} &.220 ± .026&.189 ± .026&.201 ± .028&.204 ± .041&.187 ± .028\\
        \midrule
        
        \multirow{2}{*}{\textbf{Zeno}}
        & \textbf{ADP} &.215 ± .002&.215 ± .001&.213 ± .003&.216 ± .001&.141 ± .003\\
        & \textbf{GTG} &.199 ± .028&.187 ± .044&.193 ± .044&.225 ± .031&.196 ± .038\\
        \bottomrule
    \end{tabular}}

    \vspace{0.5cm}
    \caption{CE scores with noisy IID clients.}
    \label{tab:AS_noise}
    \resizebox{0.47\textwidth}{!}{%
    \begin{tabular}{ll||c|c|c|c|c}
        \toprule
        \textbf{Method} & \textbf{Metric} & \textbf{Cli. 1} & \textbf{Cli. 2} & \textbf{Cli. 3} & \textbf{Cli. 4} & \textbf{Cli. 5} \\
        \midrule
        
        \multirow{2}{*}{\textbf{FedAvg}}
        & \textbf{ADP} &.215 ± .001&.213 ± .000&.206 ± .000&.193 ± .000&.173 ± .001\\
        & \textbf{GTG} &.203 ± .031&.193 ± .049&.197 ± .057&.218 ± .030&.189 ± .025\\
        \midrule
        
        \multirow{2}{*}{\textbf{FedNova}}
        & \textbf{ADP} &.215 ± .001&.213 ± .000&.206 ± .000&.193 ± .001&.173 ± .001\\
        & \textbf{GTG} &.215 ± .03&.214 ± .023&.190 ± .053&.180 ± .049&.200 ± .065\\
        \midrule
        
        \multirow{2}{*}{\textbf{FedProx}}
        & \textbf{ADP} &.215 ± .001&.214 ± .001&.206 ± .001&.192 ± .001&.172 ± .001\\
        & \textbf{GTG} &.182 ± .036&.200 ± .049&.191 ± .050&.226 ± .062&.201 ± .036\\
        \midrule
        
        \multirow{2}{*}{\textbf{Krum}}
        & \textbf{ADP} &.188 ± .003&.190 ± .003&.192 ± .004&.219 ± .006&.212 ± .013\\
        & \textbf{GTG} &.194 ± .047&.195 ± .031&.201 ± .033&.208 ± .03&.203 ± .025\\
        \midrule
        
        \multirow{2}{*}{\textbf{Zeno}}
        & \textbf{ADP} &.211 ± .003&.217 ± .001&.212 ± .001&.190 ± .003&.170 ± .001\\
        & \textbf{GTG} &.187 ± .072&.227 ± .049&.198 ± .100&.201 ± .097&.187 ± .098\\
        \bottomrule
    \end{tabular}}
\end{table}

\vspace{0.1cm}\subsubsection{Byzantine Fault Tolerant Aggregation}

Next, we investigated the impact of robust aggregators (Krum and Zeno) in a setting with a baseline adversary performing label- or gradient-flipping attacks (detailed in Table~\ref{tab:AS_attack}, with Client 5 as the attacker). ADP proved to be a highly effective misbehavior detection tool. Owing to its direct comparison of gradient directions, it consistently assigned the lowest score to the attacker, an effect that remained stable regardless of the BFT aggregation technique used. In contrast, the marginal difference-based GTG was highly sensitive to the aggregation mechanism. 

This is due to the coalition evaluations GTG performs: its scores are directly influenced by how the BFT aggregator filters or modifies the attacker's update. Interestingly, GTG sometimes assigned a higher score to the attacker when Krum or Zeno were used compared to FedAvg. This indicates that the robust aggregators were successful in mitigating the attacker's negative influence to the point that their (now dampened) update was not considered significantly harmful in GTG's coalition-based evaluation. This is confirmed by the statistical tests in Table~\ref{tab:AS_test}, which show that GTG scores for benign clients remain stable across aggregators, whereas ADP scores for all clients are drawn from distinct distributions.

\begin{table}[t]
    \centering
    \caption{Root Mean Square Error of the aggregation-specific scores compared to the default (FedAvg).}
    \label{tab:AS_agg}
    \resizebox{.47\textwidth}{!}{%
    \begin{tabular}{c||c|c||c!{\vrule width 2pt}c||c|c}
        \toprule
        \textbf{CE} & \textbf{Prox} & \textbf{Nova} &\multicolumn{2}{c||}{\textbf{Setting}}& \textbf{Krum} & \textbf{Zeno} \\
        \midrule
        \textbf{ADP} & $0.17$ & $0.17$ & \textbf{High} & \textbf{Label} & $0.00$ & $0.00$ \\
        \textbf{GTG} & $0.12$ & $0.11$ & \textbf{Imb.}\ & \textbf{Flip}  & $0.10$ & $0.02$ \\
        \midrule
        \textbf{ADP} & $0.00$ & $0.00$ & \textbf{Low}  & \textbf{Grad}  & $0.02$ & $0.00$ \\
        \textbf{GTG} & $0.05$ & $0.07$ & \textbf{Imb.}\ & \textbf{Flip}  & $0.01$ & $0.04$ \\
        \midrule
        \textbf{ADP} & $0.00$ & $0.01$ & \multicolumn{2}{c||}{\multirow{2}{*}{\textbf{Noise}}} & $0.03$ & $0.02$ \\
        \textbf{GTG} & $0.02$ & $0.02$ & \multicolumn{2}{c||}{}         & $0.08$ & $0.07$ \\
        \bottomrule
    \end{tabular}}

    \vspace{0.5cm}
    \caption{Anderson-Darling $p$-values of the client's scores based on FedAvg and other aggregation techniques. Statistically relevant result corresponds to value below $p\ll0.05$.}
    \label{tab:AS_test}
    \begin{adjustbox}{max width=\textwidth}
    \begin{tabular}{cc||c|c|c|c|c}
        \toprule
        \textbf{Setting} & \textbf{CE} & \textbf{1} & \textbf{2} & \textbf{3} & \textbf{4} & \textbf{5} \\
        \midrule\midrule
        \multirow{2}{*}{\textbf{High}}
          & \textbf{ADP} & 0.00 &  0.25 & 0.00 &  0.01 &  0.01 \\
          & \textbf{GTG} &   0.25 &   0.25 &   0.25 &   0.25 &   0.25 \\
        \midrule
        \multirow{2}{*}{\textbf{Low}}
          & \textbf{ADP} & 0.00 &   0.25 &   0.25 &   0.25 &   0.01 \\
          & \textbf{GTG} &   0.25 &   0.25 &   0.25 &   0.25 &   0.25 \\
        \midrule\midrule
        \multirow{2}{*}{\textbf{Label}}
          & \textbf{ADP} & 0.00 & 0.00 & 0.00 & 0.00 & 0.00 \\
          & \textbf{GTG} &   0.25 &   0.25 &   0.25 &   0.25 &   0.25 \\
        \midrule
        \multirow{2}{*}{\textbf{Grad.}}
          & \textbf{ADP} & 0.00 & 0.00 & 0.00 & 0.00 & 0.00 \\
          & \textbf{GTG} &   0.25 &   0.25 &   0.25 &   0.25 &   0.25 \\
        \midrule\midrule
        \multirow{2}{*}{\textbf{Noise}}
          & \textbf{ADP} & 0.00 & 0.00 & 0.00 & 0.00 & 0.00 \\
          & \textbf{GTG} &   0.21&   0.19&   0.25 & 0.04 &   0.11 \\
        \bottomrule
    \end{tabular}
    \end{adjustbox}
\end{table}

\vspace{0.1cm}\subsubsection{Controlled Data Quality}

Finally, we analyzed the system's response to linearly increasing label noise (Table~\ref{tab:AS_noise}) to understand how methods handle graded data quality issues. GTG's scoring appeared arbitrary and did not strictly follow the noise ratios, suggesting a more complex interaction between data corruption and perceived utility. On the other hand, ADP performed well in all scenarios except with Krum, consistently scoring clients in inverse proportion to their noise level. We suspect that this exception arises from Krum's difficulty in selecting the ``lesser evil,'' as nearly all clients exhibited some degree of malicious behavior.

\subsection{Intentional Manipulation}

\vspace{0.1cm}\subsubsection{Self Improvement}

\begin{figure}[t]
\centering
    \caption{Clients' LOO and GTG scores in the third training round when ADULT is split into 5 clients (IID or non-IID) where the 1st client either self improving its score or not. Blue is the baseline no attack; red is attack.}
    \label{fig:SI}
    \begin{subfigure}{0.24\textwidth}
        \caption{LOO/IID. }
        \label{fig:SI_L_A_I}
        \includegraphics[width=\linewidth]{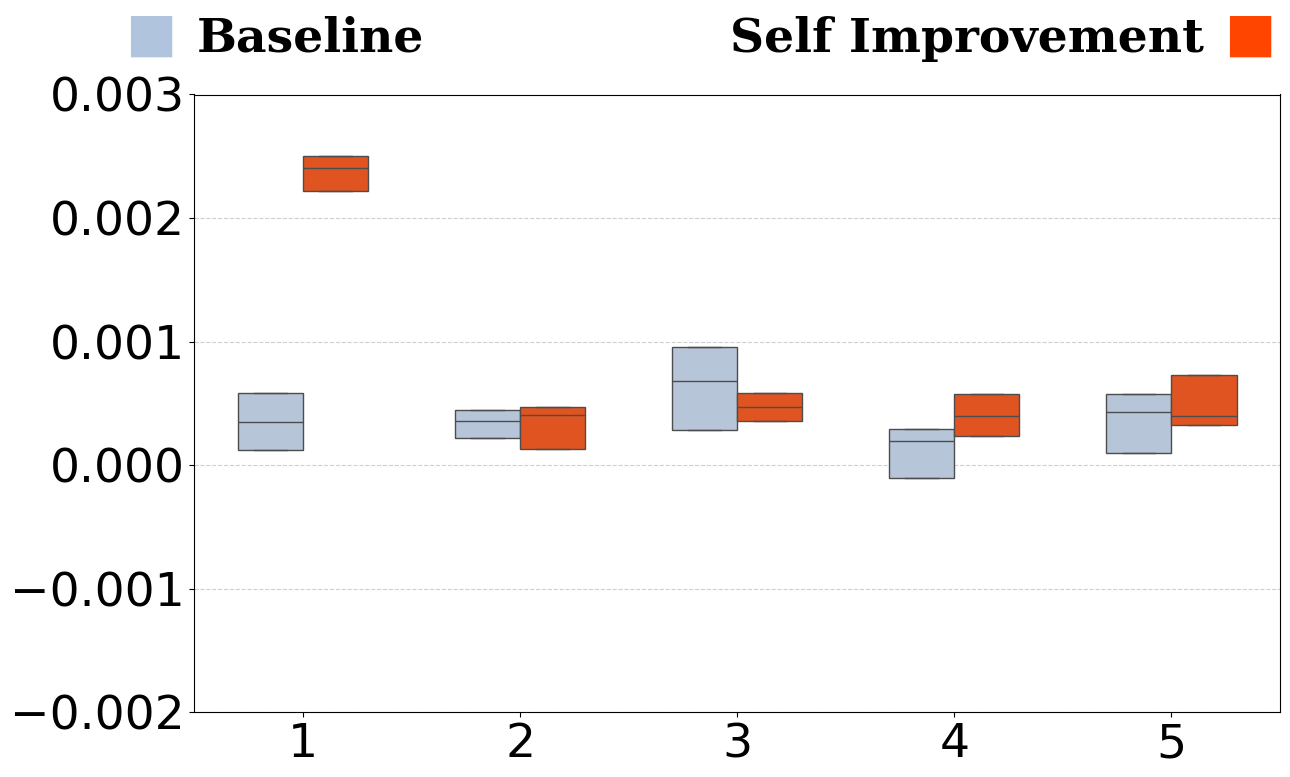}
    \end{subfigure}
    \hfill
    \begin{subfigure}{0.24\textwidth}
        \caption{LOO/non-IID. }
        \label{fig:SI_L_A_N}
        \includegraphics[width=\linewidth]{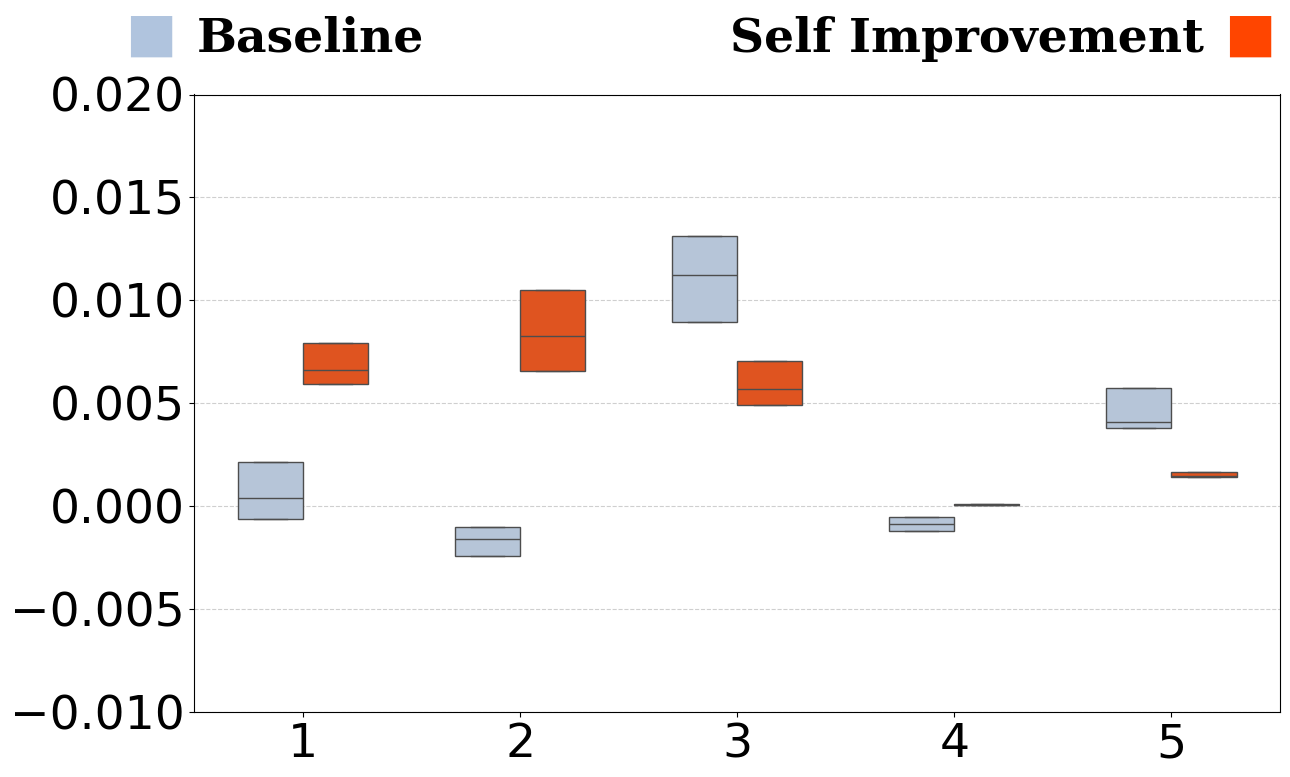}
    \end{subfigure}
    \hfill
    \begin{subfigure}{0.24\textwidth}
        \caption{GTG/IID. }
        \label{fig:SI_G_A_I}
        \includegraphics[width=\linewidth]{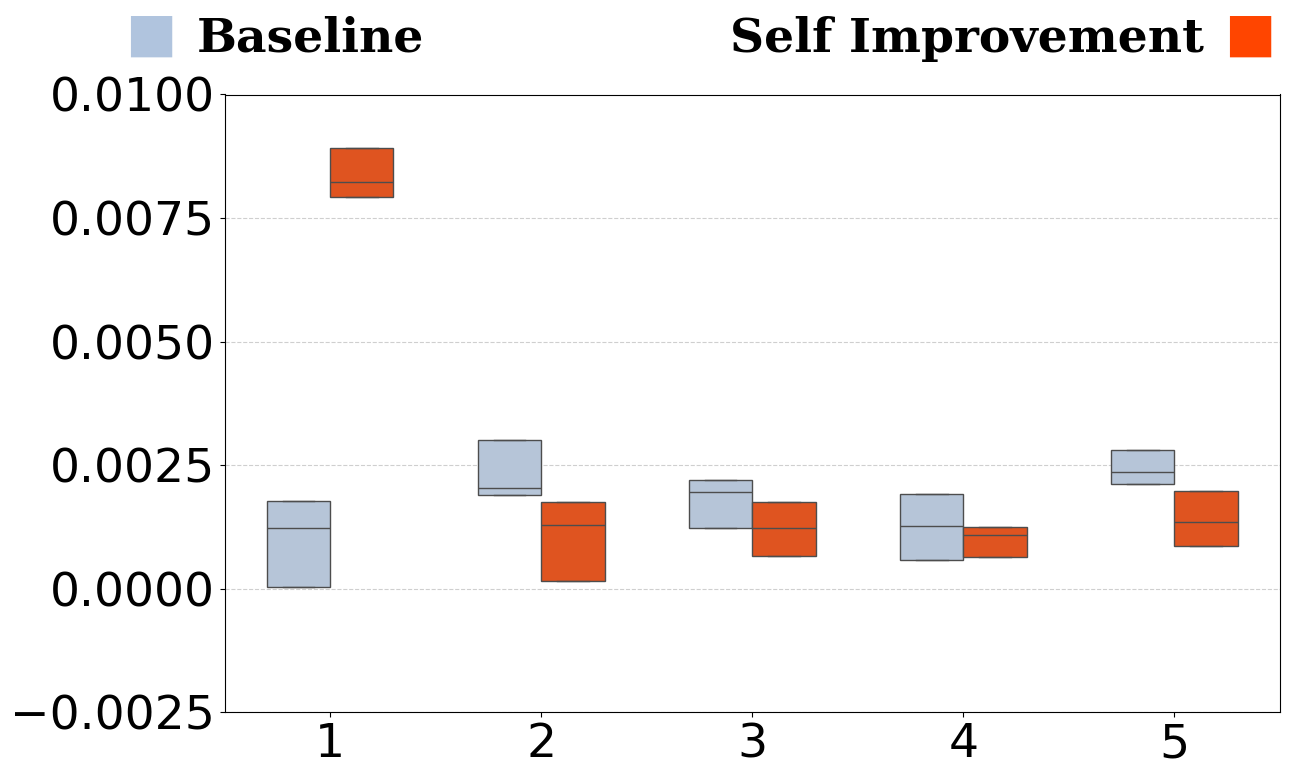}
    \end{subfigure}
    \hfill
    \begin{subfigure}{0.24\textwidth}
        \caption{GTG/non-IID. }
        \label{fig:SI_G_A_N}
        \includegraphics[width=\linewidth]{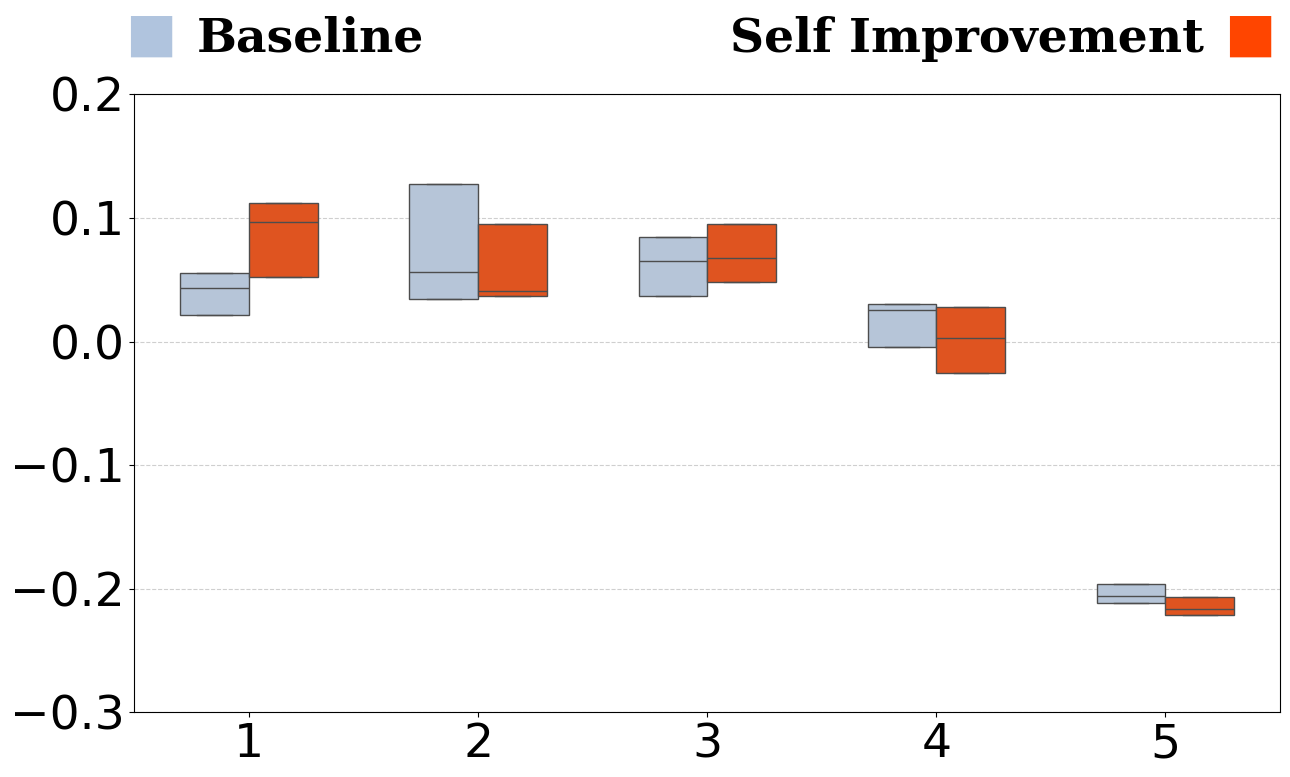}
    \end{subfigure}
\end{figure}

In this scenario, the first client acts as the attacker aiming to maximize its own score on the ADULT dataset. Figure~\ref{fig:SI} shows the resulting client contribution scores during the third training round, with LOO scores on the top (Figures~\ref{fig:SI_L_A_I} and~\ref{fig:SI_L_A_N}) and GTG on the bottom (Figures~\ref{fig:SI_G_A_I} and~\ref{fig:SI_G_A_N}), under both IID (left-side) and non-IID (right-side) conditions. The results clearly indicate that the \emph{Self Improvement} attack is effective. The effect is especially pronounced under IID conditions (Figures~\ref{fig:SI_L_A_I} and~\ref{fig:SI_G_A_I}), where the attacker's manipulated score rises sharply with minimal impact on other clients. In contrast, under non-IID splits (Figures~\ref{fig:SI_L_A_N} and~\ref{fig:SI_G_A_N}), the attack still improves the attacker’s score but causes varied side effects on the other clients.

\begin{figure}[!t]
\centering
    \caption{Clients' contribution score differences (LOO (blue) or GTG (red)) in the third training round when ADULT is split into 5 clients (IID or non-IID) when the 1st client is self improving its score.}
    \label{fig:self_diff}
    \begin{subfigure}{0.24\textwidth}
        \caption{IID. }
        \label{fig:self_diff_A_I}
        \includegraphics[width=\linewidth]{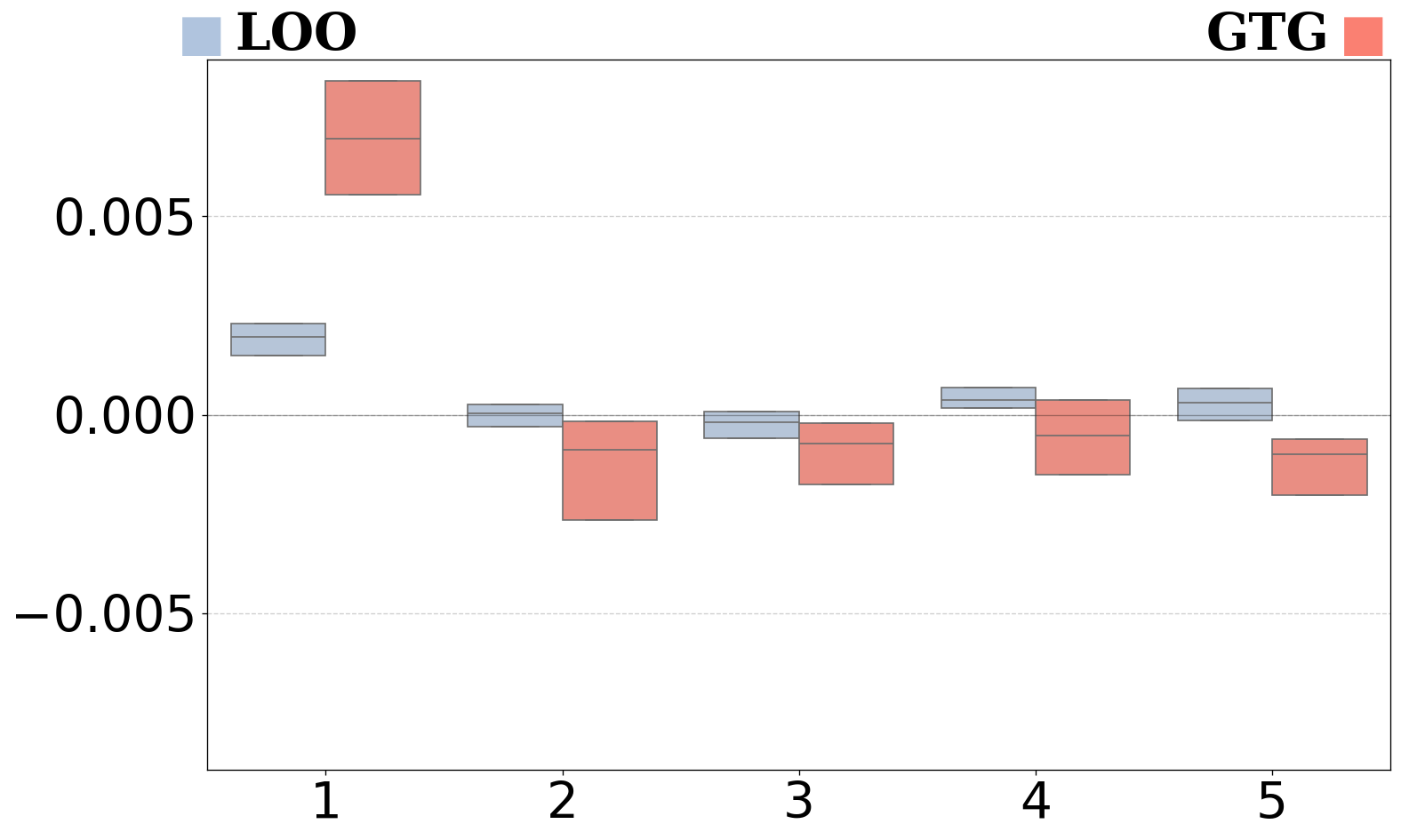}
    \end{subfigure}
    \hfill
    \begin{subfigure}{0.24\textwidth}
        \caption{non-IID. }
        \label{fig:self_diff_A_N}
        \includegraphics[width=\linewidth]{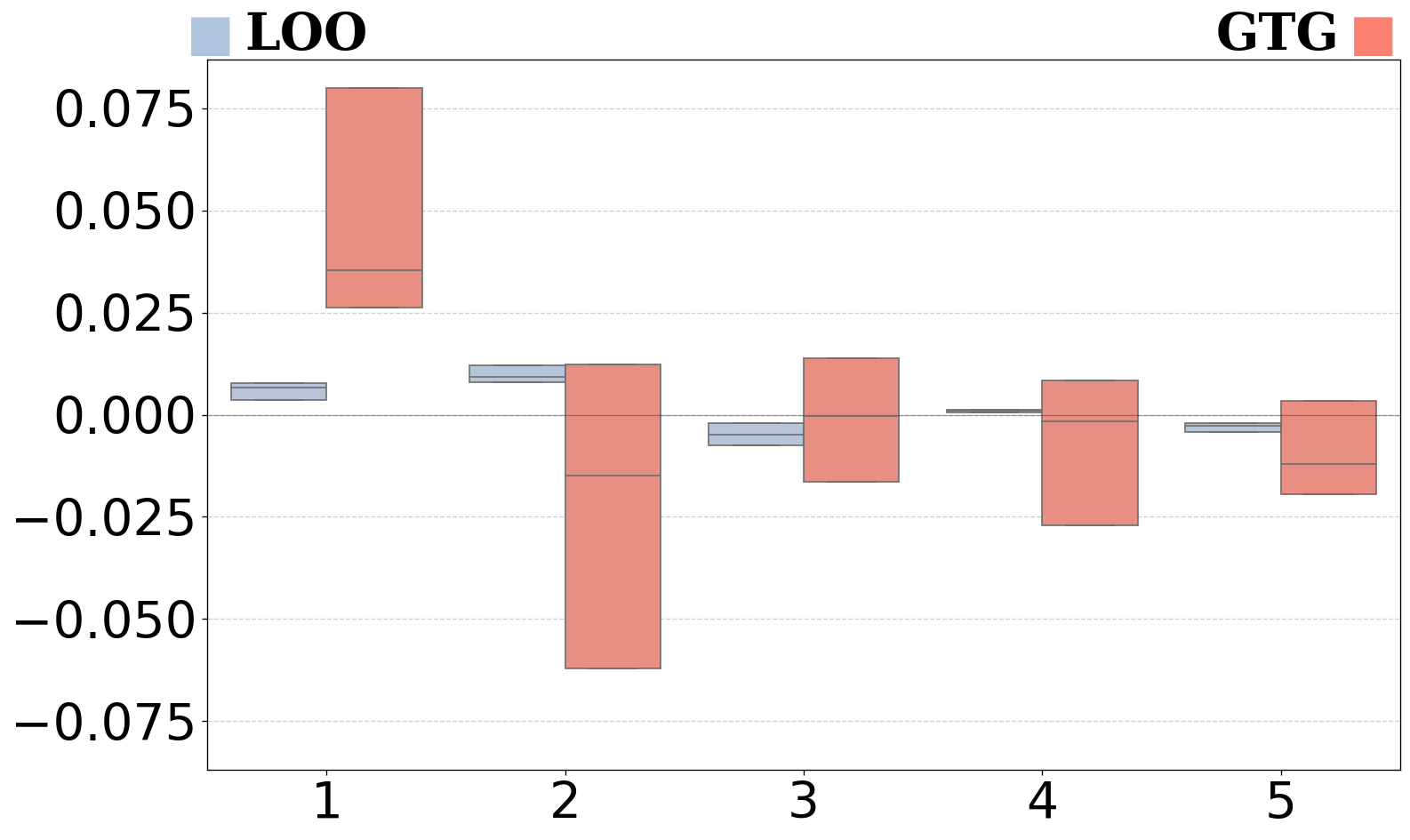}
    \end{subfigure}
\end{figure}

Figure~\ref{fig:self_diff} provides a closer look at the score changes, revealing different sensitivities between the CE schemes. The score differences for LOO are generally smaller, and close to zero in the non-IID case (\ref{fig:self_diff_A_N}), which is expected since LOO compares coalitions with larger cardinality, limiting individual influence. GTG, on the other hand, is more sensitive due to its focus on smaller coalitions; as shown in the IID case (\ref{fig:self_diff_A_I}), the attacker's score increase is often accompanied by a decrease in the scores of other clients.

In Table~\ref{tab:SI_diff}, we analyze the absolute and relative changes to the attacker's score across multiple rounds. A surprising pattern emerges: although the attack has a smaller absolute effect on LOO, the relative change is often more pronounced compared to GTG. This is due to LOO's reliance on a single marginal contribution, which has a direct impact on the scores. In contrast, GTG aggregates multiple marginal contributions; therefore, the manipulation has a more diluted relative effect.

To validate these findings, we conducted a one-sided paired $t$-test to see whether the scores indeed increased. The results in Table~\ref{tab:SI_p} show that the score increases are statistically significant ($p \ll 0.05$) in all but one case (GTG in round 2, non-IID). This provides strong statistical evidence that our \emph{Self Improvement} attack is effective.

\begin{table}[t]
\centering
    \caption{The attacker's (client 1) absolute ($|f(a)-f(1)|$) and relative (\gray{$\frac{|f(a)-f(1)|}{f(1)}$}) contribution score difference between Self Improve and baseline for specific rounds separately and executed altogether for the ADULT dataset.}
    \label{tab:SI_diff}
    \begin{adjustbox}{width=0.9\width,center}
    \begin{tabular}{cc|cccc}
        \multicolumn{2}{c|}{$\times10^{-2}$} & \textbf{R2} & \textbf{R3} & \textbf{R4} & \textbf{All} \\
        \midrule
        \multirow{4}{*}{\rotatebox{90}{\textbf{IID}}} & \multirow{2}{*}{\textbf{LOO}} & $0.15 \pm 0.08$ & $0.08 \pm 0.04$ & $0.06 \pm 0.03$ & $0.37 \pm 0.05$ \\
        &  & \gray{$590\% \pm 598$} & \gray{$214\% \pm 129$} & \gray{$155\% \pm 101$} & \gray{$2120\% \pm 1556$} \\
        \cline{2-6}
        & \multirow{2}{*}{\textbf{GTG}} & $0.49 \pm 0.12$ & $0.23 \pm 0.12$ & $0.15 \pm 0.04$ & $0.44 \pm 0.11$ \\
        &  & \gray{$160\% \pm 61$} & \gray{$103\% \pm 70$} & \gray{$76\% \pm 29$} & \gray{$236\% \pm 80$} \\
        \midrule
        \multirow{4}{*}{\rotatebox{90}{\textbf{non-IID}}} & \multirow{2}{*}{\textbf{LOO}} & $0.28 \pm 0.18$ & $0.16 \pm 0.17$ & $0.08 \pm 0.06$ & $0.78 \pm 0.17$ \\
        &  & \gray{$635\% \pm 125$} & \gray{$210\% \pm 189$} & \gray{$207\% \pm 386$} & \gray{$3862\% \pm 762$} \\
        \cline{2-6}
        & \multirow{2}{*}{\textbf{GTG}} & $2.75 \pm 2.07$ & $2.37 \pm 2.04$ & $1.97 \pm 1.23$ & $1.98 \pm 1.31$ \\
        &  & \gray{$68\% \pm 60$} & \gray{$58\% \pm 61$} & \gray{$54\% \pm 62$} & \gray{$39\% \pm 30$} \\
        \midrule
    \end{tabular}
    \end{adjustbox}

    \vspace{0.5cm}
    \caption{The $p$-values for a paired one-sided $t$-test comparing the Self Improve and baseline scores across various rounds separately and together for ADULT. Statistically relevant result corresponds to value below $p\ll0.05$.}
    \label{tab:SI_p}
    \begin{tabular}{cc|cccc}
        && \textbf{R2} & \textbf{R3} & \textbf{R4} & \textbf{All} \\
        \midrule
        \multirow{2}{*}{\textbf{IID}} & \textbf{LOO} & 0.00 & 0.00 & 0.01 & 0.00 \\
        & \textbf{GTG} & 0.00 & 0.00 & 0.00 & 0.00 \\
        \midrule
        \multirow{2}{*}{\textbf{nonn-IID}} & \textbf{LOO} & 0.01 & 0.01 & 0.00 & 0.00 \\
        & \textbf{GTG} & 0.08 & 0.01 & 0.02 & 0.00 \\
        \midrule
    \end{tabular}
\end{table}

\vspace{0.1cm}\subsubsection{Targeted Decrease}

\begin{table}[t]
\centering
    \caption{The targeted client's (client 2) absolute ($|f(\hat{2})-f(2)|$) and relative (\gray{$\frac{|f(\hat{2})-f(2)|}{f(2)}$}) contribution score difference between Targeted Decrease and the baseline for selective round separately and together.}
    \label{tab:TD_diff}
    \begin{adjustbox}{width=0.75\width,center}
    \begin{tabular}{ccc|cccc}
        \multicolumn{3}{c|}{$\times10^{-2}$ (\gray{$\times100$})} & \textbf{R2} & \textbf{R3} & \textbf{R4} & \textbf{All} \\
        \midrule
        \multirow{8}{*}{\rotatebox{90}{\textbf{ADULT}}} & \multirow{4}{*}{\rotatebox{90}{\textbf{IID}}} & \multirow{2}{*}{LOO} & $0.06 \pm 0.06$ & $0.14 \pm 0.25$ & $0.04 \pm 0.02$ & $0.21 \pm 0.20$ \\
        & &  & \gray{$1.9\% \pm 2.8$} & \gray{$4.6\% \pm 7.9$} & \gray{$1.7\% \pm 2.0$} & \gray{$5.2\% \pm 3.6$} \\
        \cline{3-7}
        & & \multirow{2}{*}{\textbf{GTG}} & $0.29 \pm 0.25$ & $0.11 \pm 0.07$ & $0.05 \pm 0.03$ & $0.23 \pm 0.21$ \\
        & &  & \gray{$1.0\% \pm 1.1$} & \gray{$0.4\% \pm 0.3$} & \gray{$0.2\% \pm 0.1$} & \gray{$1.7\% \pm 1.7$} \\
        \cline{2-7}
        & \multirow{4}{*}{\rotatebox{90}{\textbf{non-IID}}} & \multirow{2}{*}{\textbf{LOO}} & $4.97 \pm 1.01$ & $2.75 \pm 0.64$ & $1.92 \pm 0.25$ & $25.79 \pm 2.57$ \\
        & &  & \gray{$165\% \pm 215$} & \gray{$614\% \pm 514$} & \gray{$126\% \pm 276$} & \gray{$263\% \pm 168$} \\
        \cline{3-7}
        & & \multirow{2}{*}{\textbf{GTG}} & $7.07 \pm 4.48$ & $4.04 \pm 2.35$ & $3.16 \pm 2.20$ & $5.25 \pm 3.25$ \\
        & &  & \gray{$1.2\% \pm 0.7$} & \gray{$0.6\% \pm 0.2$} & \gray{$0.5\% \pm 0.3$} & \gray{$0.9\% \pm 0.4$} \\
        \midrule
        \multirow{8}{*}{\rotatebox{90}{\textbf{Fashion}}} & \multirow{4}{*}{\rotatebox{90}{\textbf{IID}}} & \multirow{2}{*}{\textbf{LOO}} & $12.17 \pm 2.65$ & $6.28 \pm 1.24$ & $4.33 \pm 1.25$ & $41.45 \pm 5.28$ \\
        & &  & \gray{$68\% \pm 58$} & \gray{$105\% \pm 267$} & \gray{$14\% \pm 167$} & \gray{$174\% \pm 139$} \\
        \cline{3-7}
        & & \multirow{2}{*}{\textbf{GTG}} & $46.79 \pm 45.22$ & $29.71 \pm 41.58$ & $16.89 \pm 14.99$ & $108.32 \pm 124.39$ \\
        & &  & \gray{$31\% \pm 34$} & \gray{$23\% \pm 26$} & \gray{$17\% \pm 21$} & \gray{$965\% \pm 1160$} \\
        \cline{2-7}
        & \multirow{4}{*}{\rotatebox{90}{\textbf{non-IID}}} & \multirow{2}{*}{\textbf{LOO}} & $25.77 \pm 4.05$ & $8.30 \pm 1.71$ & $5.65 \pm 1.50$ & $86.14 \pm 12.04$ \\
        & &  & \gray{$34\% \pm 50$} & \gray{$5\% \pm 2.6$} & \gray{$3\% \pm 0.8$} & \gray{$63\% \pm 37$} \\
        \cline{3-7}
        & & \multirow{2}{*}{\textbf{GTG}} & $91.52 \pm 49.94$ & $20.06 \pm 10.98$ & $11.61 \pm 4.90$ & $144.26 \pm 64.53$ \\
        & &  & \gray{$12\% \pm 10$} & \gray{$3.3\% \pm 2.7$} & \gray{$2.1\% \pm 1.1$} & \gray{$36\% \pm 19$} \\
        \midrule
    \end{tabular}
    \end{adjustbox}

    \vspace{0.5cm}
    \caption{The $p$-values for a paired two-sided $t$-test comparing Targeted Decrease with the baseline across various rounds separately and together for ADULT and FMNIST. Statistically relevant result corresponds to value below $p\ll0.05$.}
    \label{tab:TD_p}
    \begin{tabular}{ccc|cccc}
        &&& \textbf{R2} & \textbf{R3} & \textbf{R4} & \textbf{All} \\
        \midrule
        \multirow{4}{*}{\rotatebox{90}{\textbf{ADULT}}} & \multirow{2}{*}{\rotatebox{90}{\textbf{IID}}} & \textbf{LOO} & $0.05$ & $0.29$ & $0.50$  & $0.83$ \\
        & & \textbf{GTG} & $0.02$ & $0.75$ & $0.88$  & $0.01$ \\
        \cline{2-7}
        & \multirow{2}{*}{\rotatebox{90}{\textbf{nIID}}} & \textbf{LOO} & $0.00$ & $0.00$ & $0.00$  & $0.00$ \\
        & & \textbf{GTG} & $0.01$ & $0.00$ & $0.02$  & $0.01$ \\
        \midrule
        \multirow{4}{*}{\rotatebox{90}{\textbf{Fashion}}} & \multirow{2}{*}{\rotatebox{90}{\textbf{IID}}} & \textbf{LOO} & $0.00$ & $0.00$ & $0.00$  & $0.00$ \\
        & & \textbf{GTG} & $0.01$ & $0.06$ & $0.01$  & $0.01$ \\
        \cline{2-7}
        & \multirow{2}{*}{\rotatebox{90}{\textbf{nIID}}} & \textbf{LOO} & $0.00$ & $0.00$ & $0.00$  & $0.00$ \\
        & & \textbf{GTG} & $0.00$ & $0.00$ & $0.00$  & $0.00$ \\
        \midrule
    \end{tabular}
\end{table}

\begin{figure*}[t]
\centering
    \caption{Clients' contribution scores (LOO or GTG) in the third training round when the dataset (ADULT or FMNIST) is split into 5 clients (IID or non-IID) where the 1st client either decreases the 2nd client's score or not. Blue is the baseline no attack; red is attack.}
    \label{fig:TD}
    \begin{subfigure}{0.24\textwidth}
        \caption{LOO/ADULT/IID. }
        \label{fig:TD_L_A_I}
        \includegraphics[width=\linewidth]{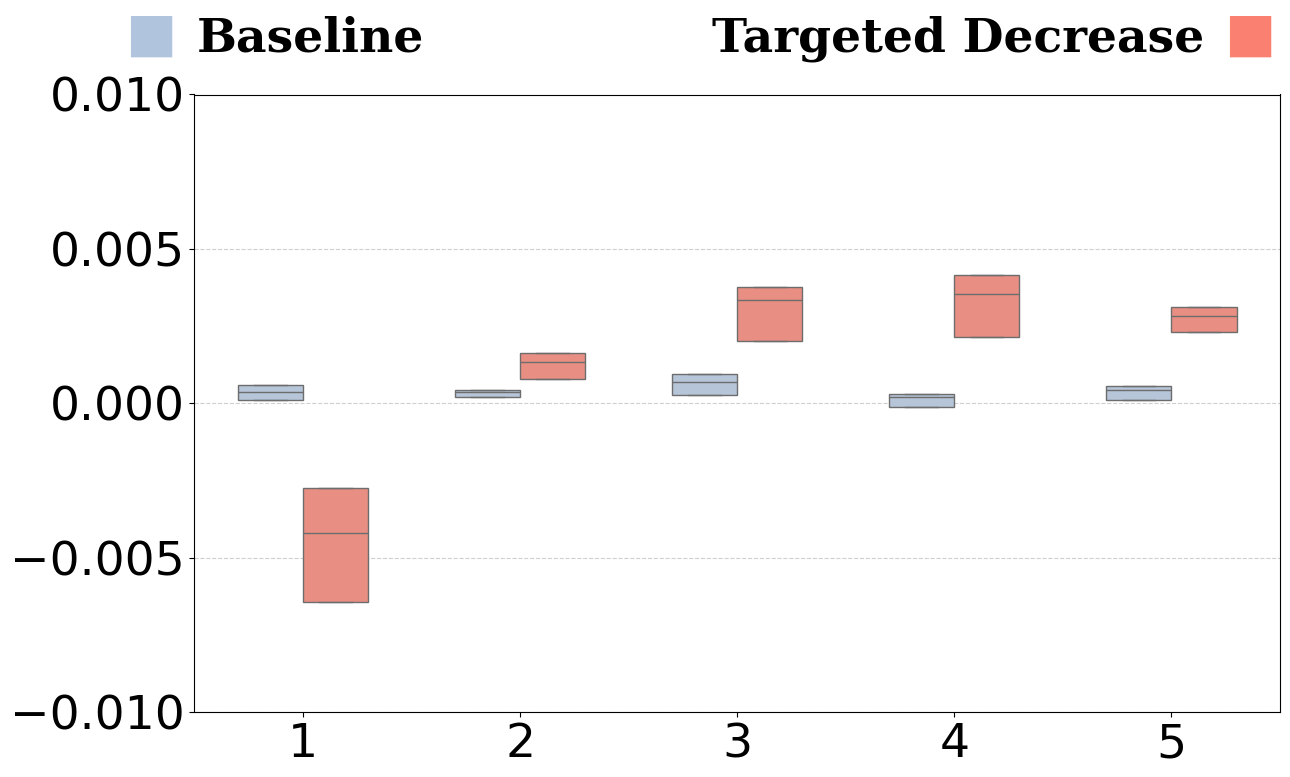}
    \end{subfigure}
    \hfill
    \begin{subfigure}{0.24\textwidth}
        \caption{LOO/ADULT/non-IID. }
        \label{fig:TD_L_A_N}
        \includegraphics[width=\linewidth]{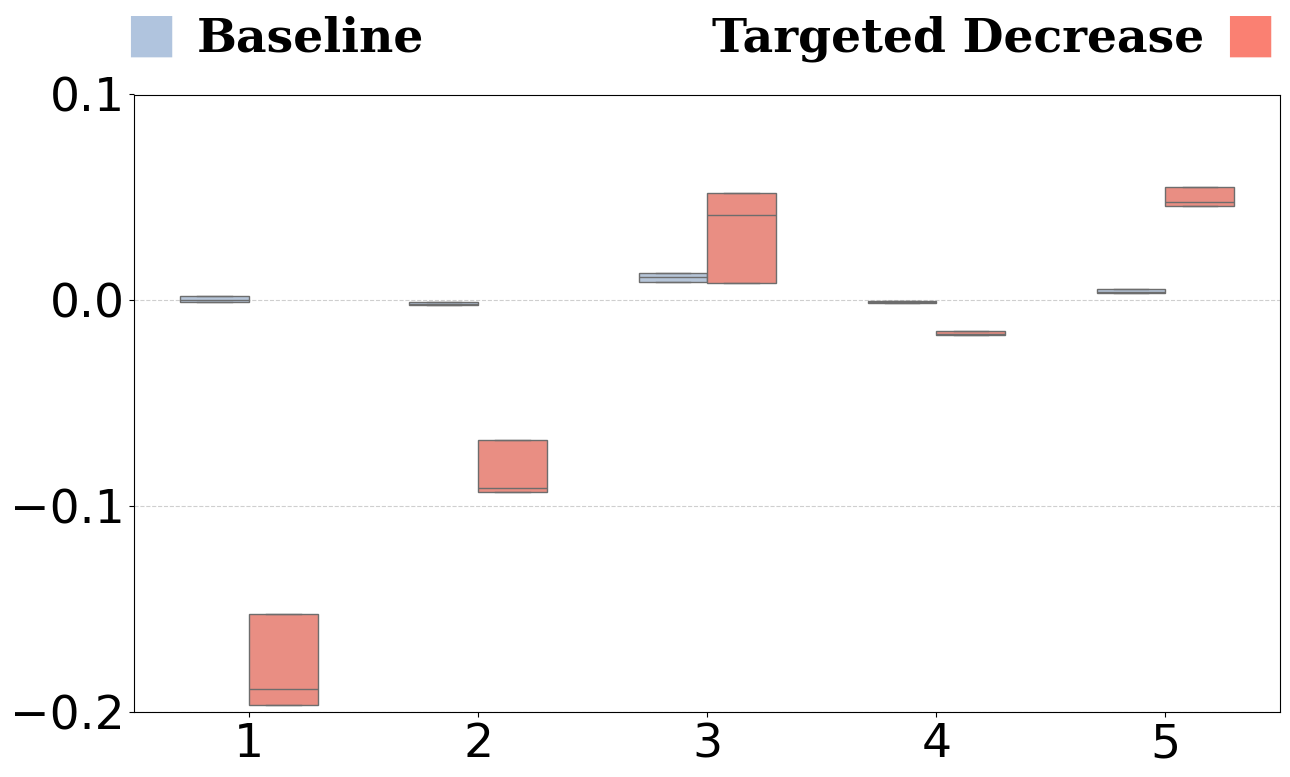}
    \end{subfigure}
    \hfill
    \begin{subfigure}{0.24\textwidth}
        \caption{LOO/Fashion/IID. }
        \label{fig:TD_L_C_I}
        \includegraphics[width=\linewidth]{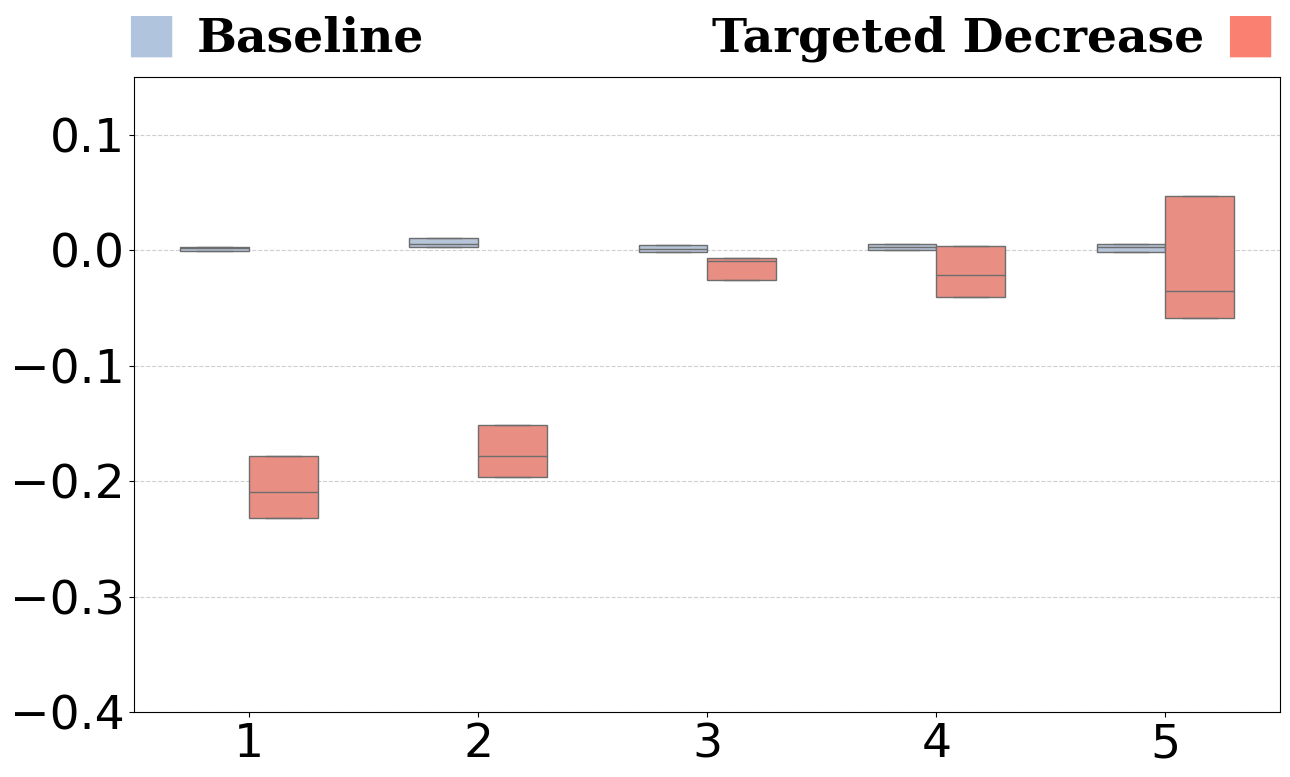}
    \end{subfigure}
    \hfill
    \begin{subfigure}{0.24\textwidth}
        \caption{LOO/Fashion/non-IID. }
        \label{fig:TD_L_C_N}
        \includegraphics[width=\linewidth]{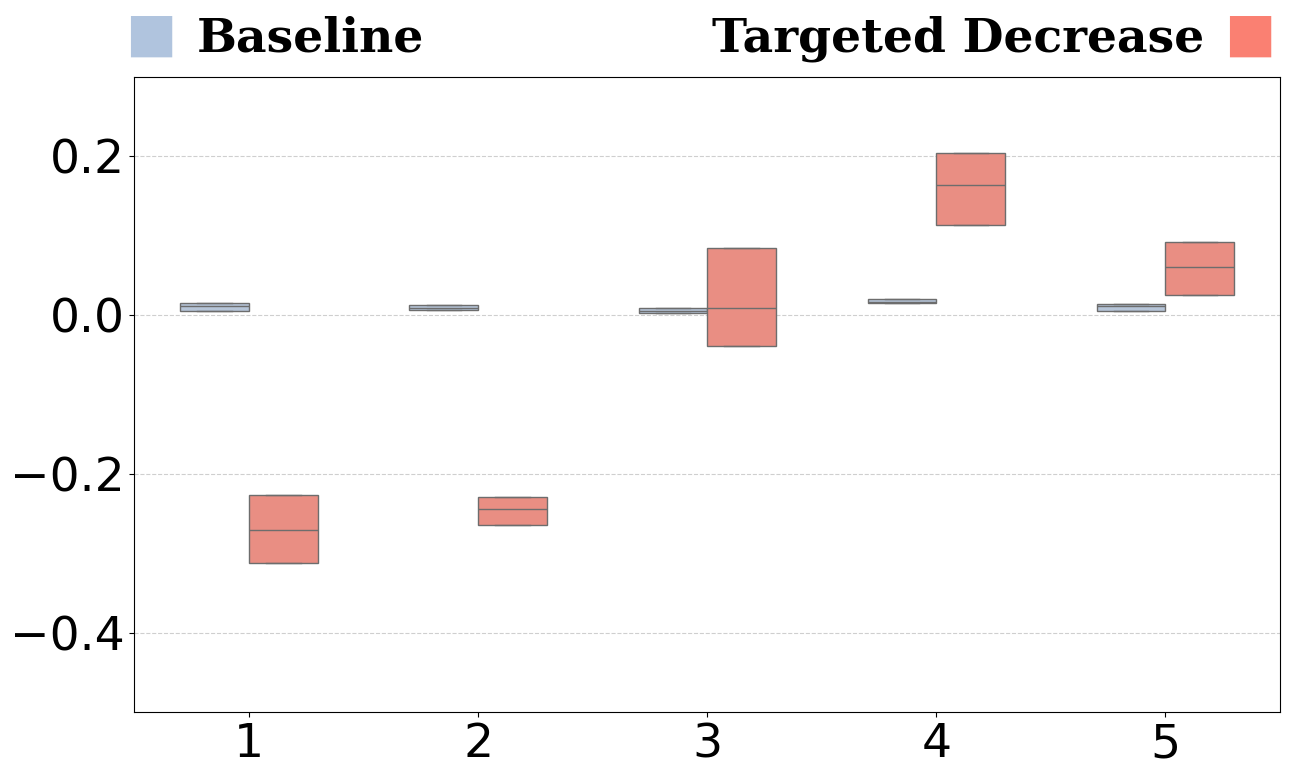}
    \end{subfigure}
    \hfill
    \begin{subfigure}{0.24\textwidth}
        \caption{GTG/ADULT/IID. }
        \label{fig:TD_G_A_I}
        \includegraphics[width=\linewidth]{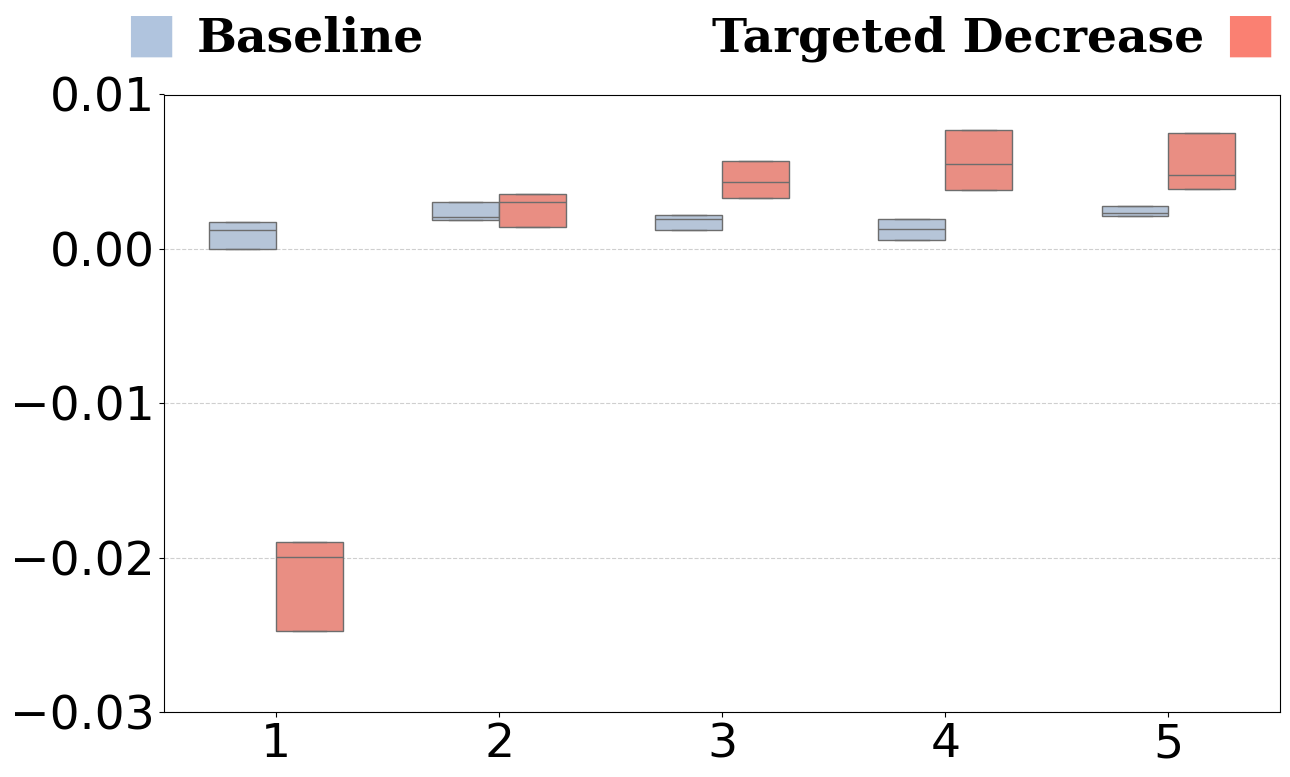}
    \end{subfigure}
    \hfill
    \begin{subfigure}{0.24\textwidth}
        \caption{GTG/ADULT/non-IID. }
        \label{fig:TD_G_A_N}
        \includegraphics[width=\linewidth]{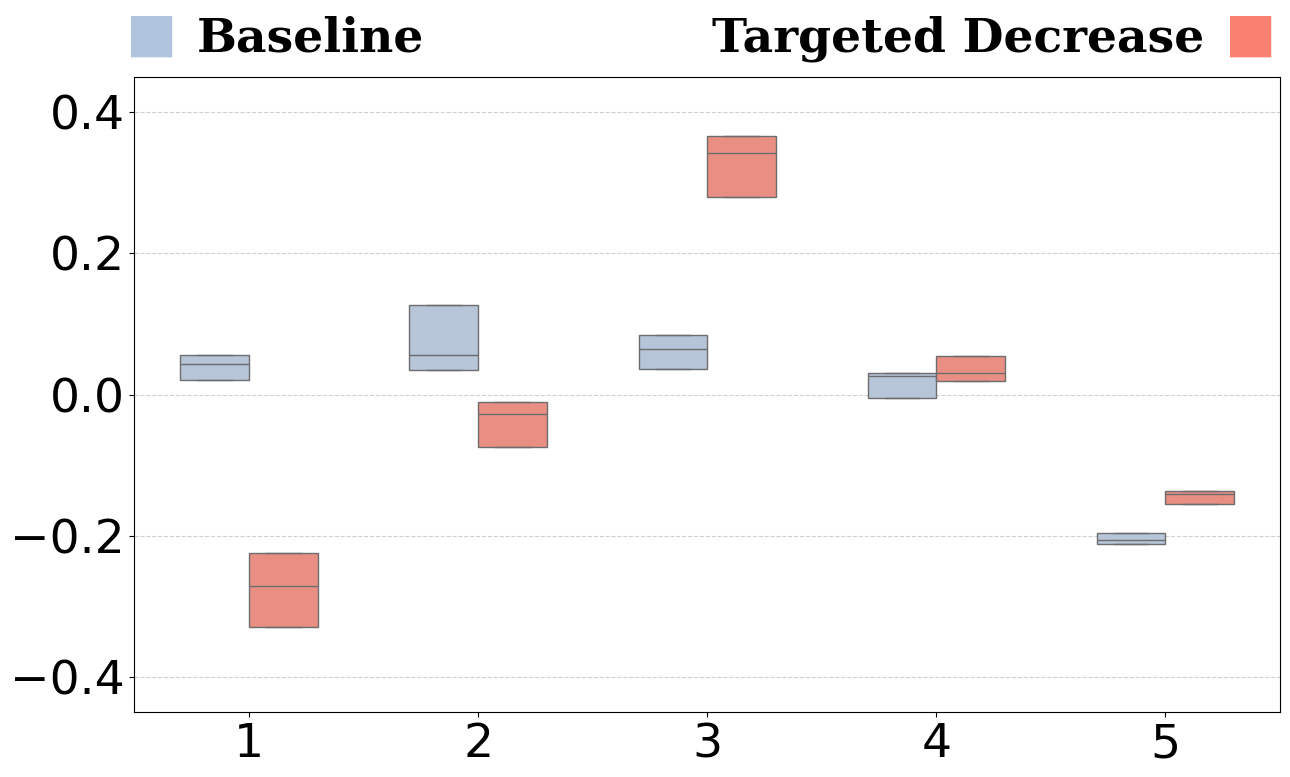}
    \end{subfigure}
    \hfill
    \begin{subfigure}{0.24\textwidth}
        \caption{GTG/Fashion/IID. }
        \label{fig:TD_G_C_I}
        \includegraphics[width=\linewidth]{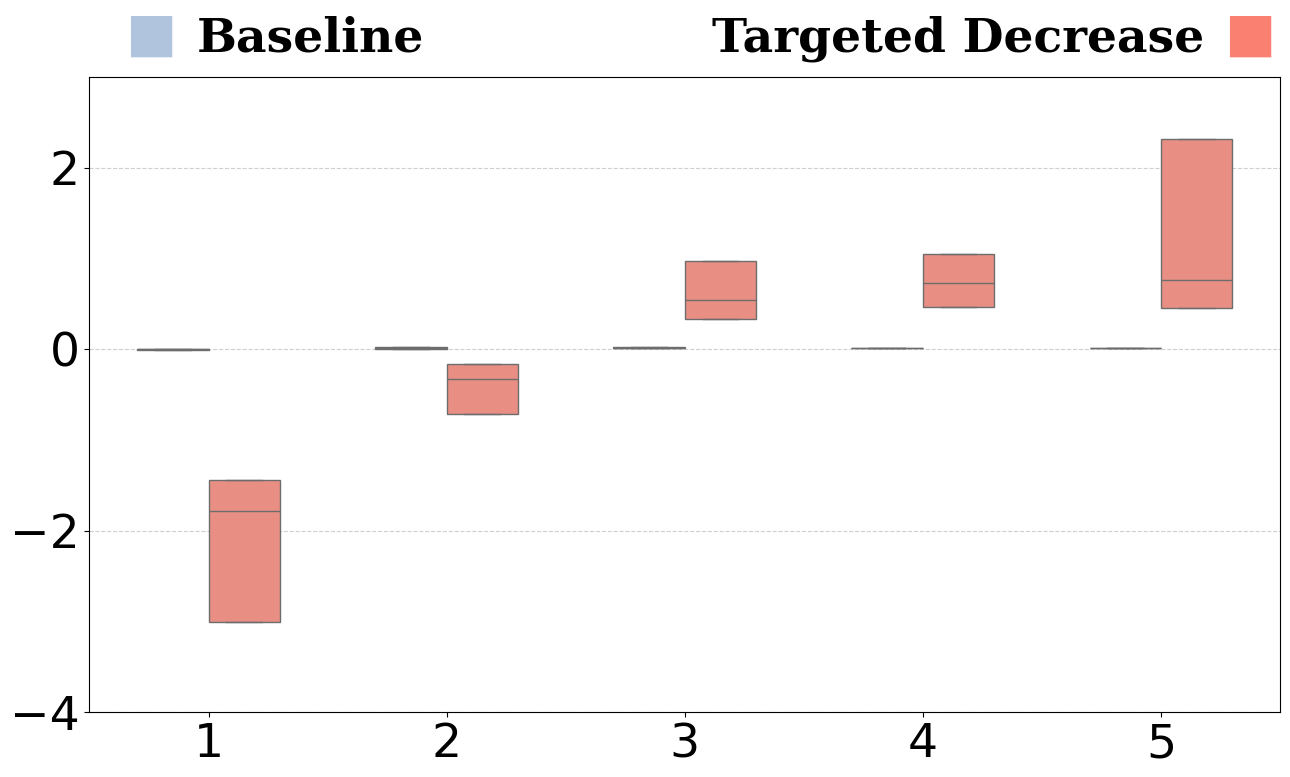}
    \end{subfigure}
    \hfill
    \begin{subfigure}{0.24\textwidth}
        \caption{GTG/Fashion/non-IID. }
        \label{fig:TD_G_C_N}
        \includegraphics[width=\linewidth]{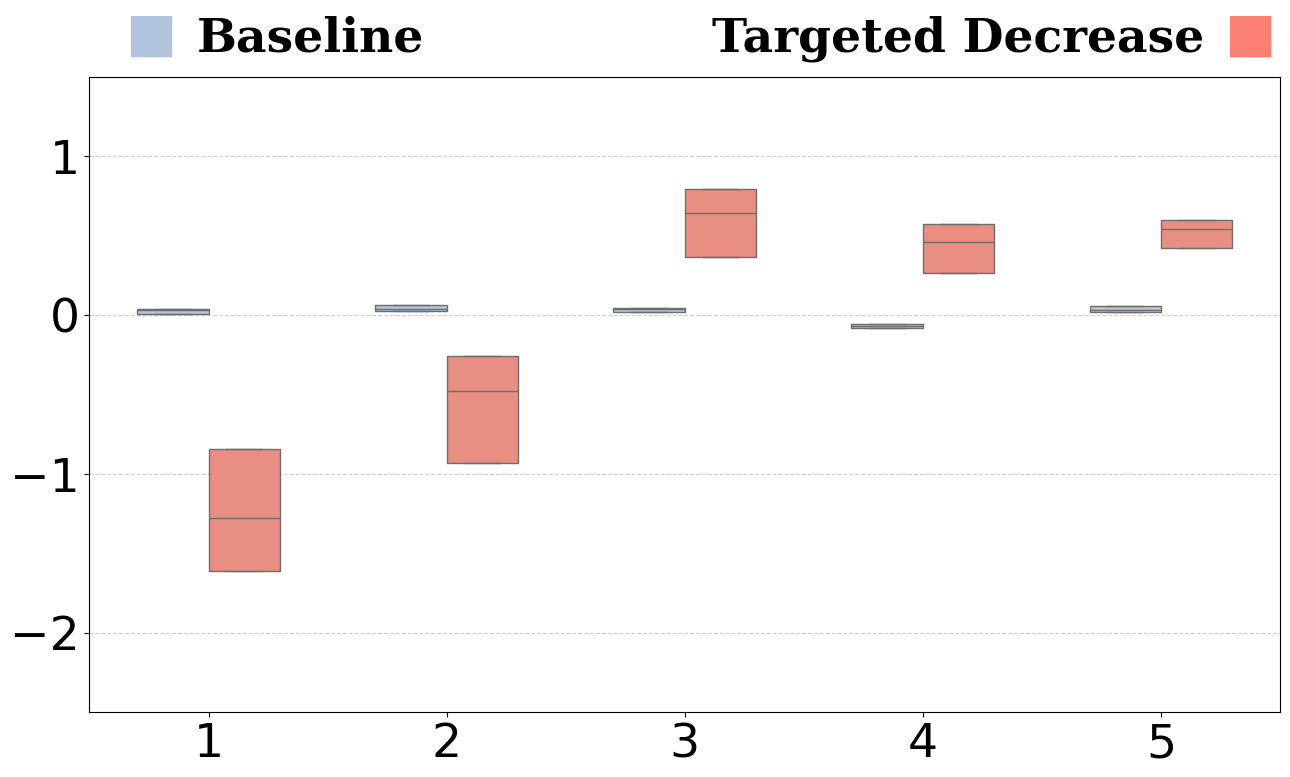}
    \end{subfigure}
\end{figure*}

\begin{figure}[t]
\centering
    \caption{Clients' contribution score differences (LOO (blue) or GTG (red)) in the third training round when the dataset (ADULT or FMNIST) is split into 5 clients (IID or non-IID) when the 1st client is decreasing the 2nd's score.}
    \label{fig:TD_diff}
    \begin{subfigure}{0.24\textwidth}
        \caption{ADULT/IID. }
        \label{fig:TD_diff_A_I}
        \includegraphics[width=\linewidth]{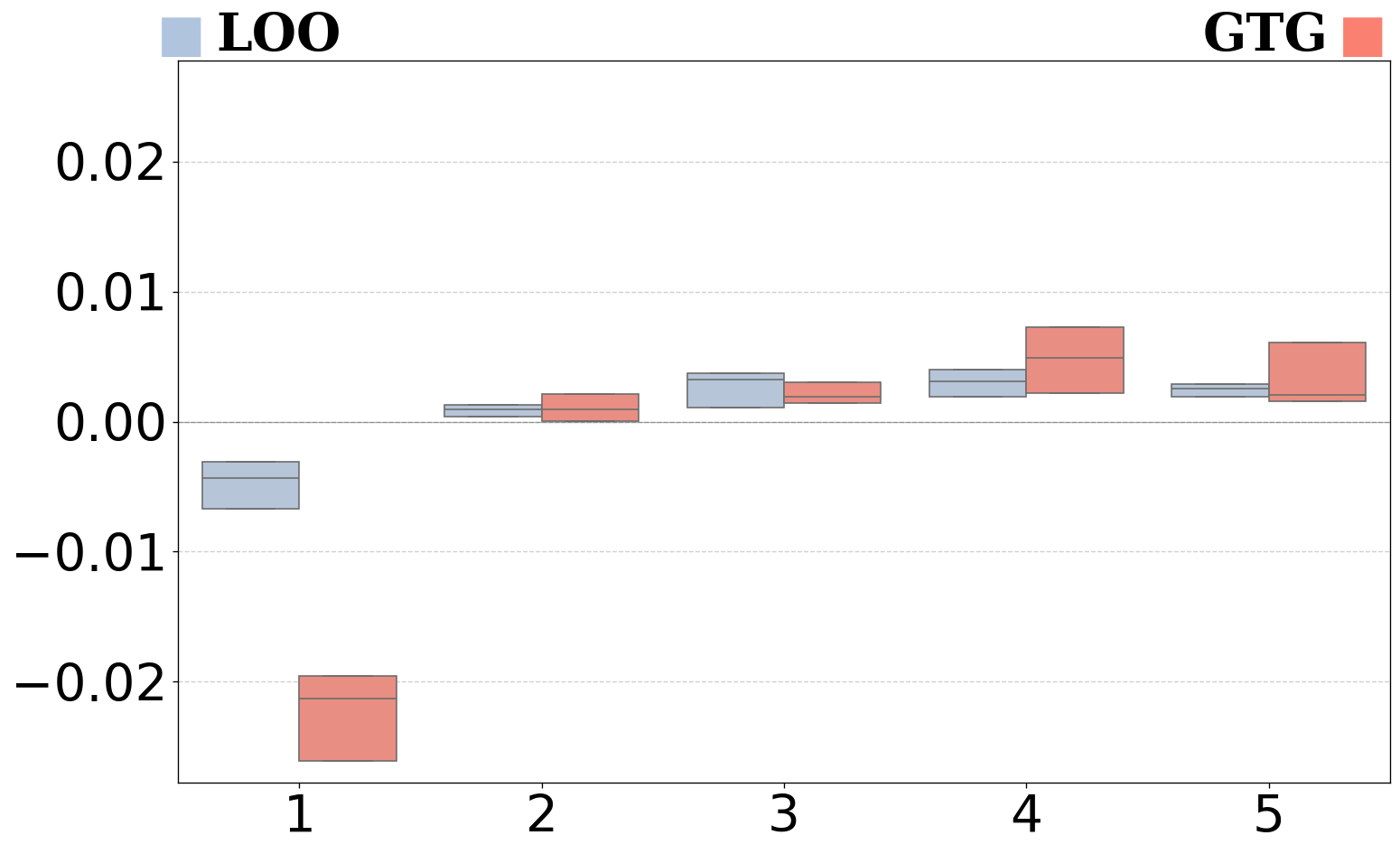}
    \end{subfigure}
    \hfill
    \begin{subfigure}{0.24\textwidth}
        \caption{ADULT/non-IID. }
        \label{fig:TD_diff_A_N}
        \includegraphics[width=\linewidth]{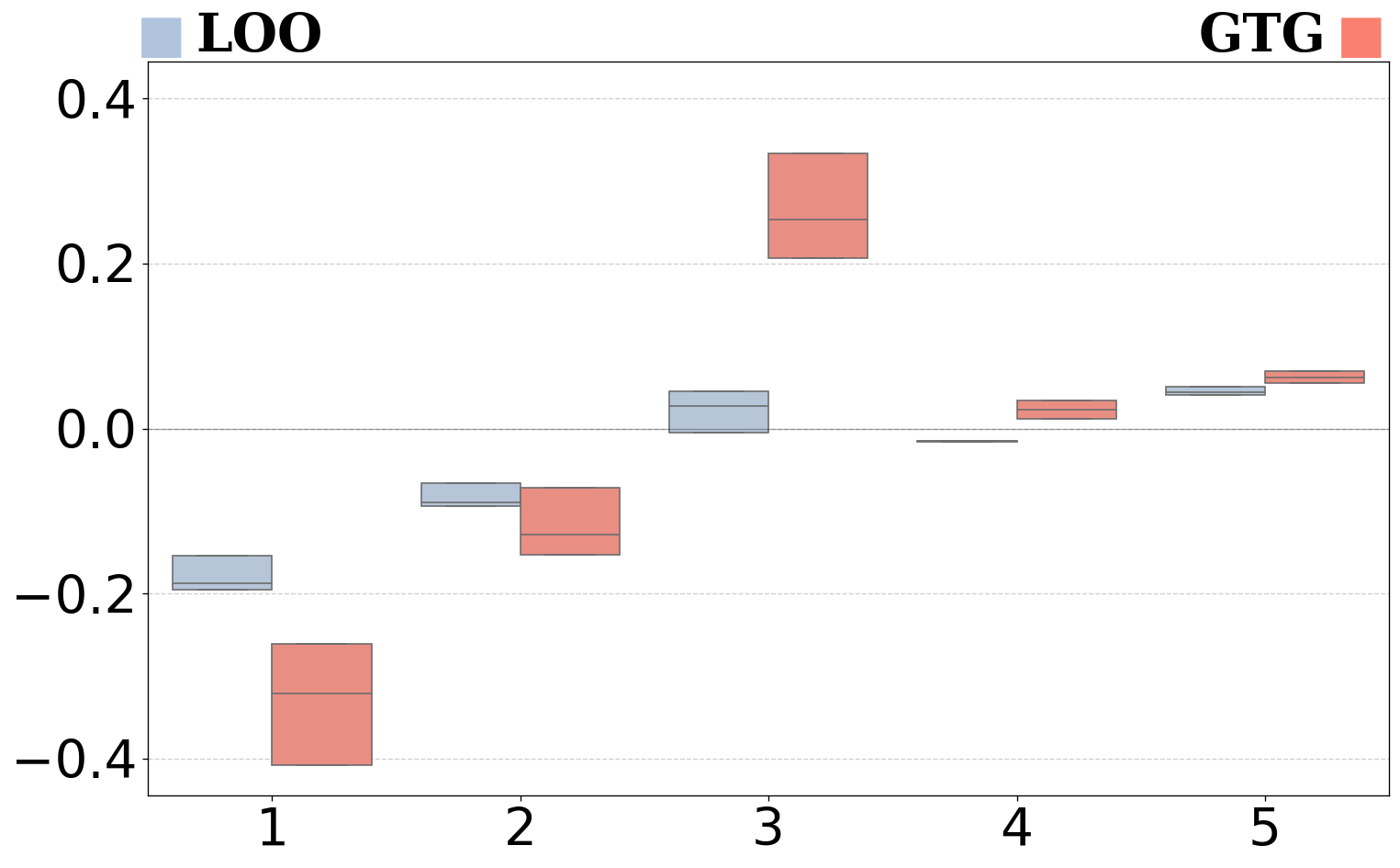}
    \end{subfigure}
    \hfill
    \begin{subfigure}{0.24\textwidth}
        \caption{Fashion/IID. }
        \label{fig:TD_diff_C_I}
        \includegraphics[width=\linewidth]{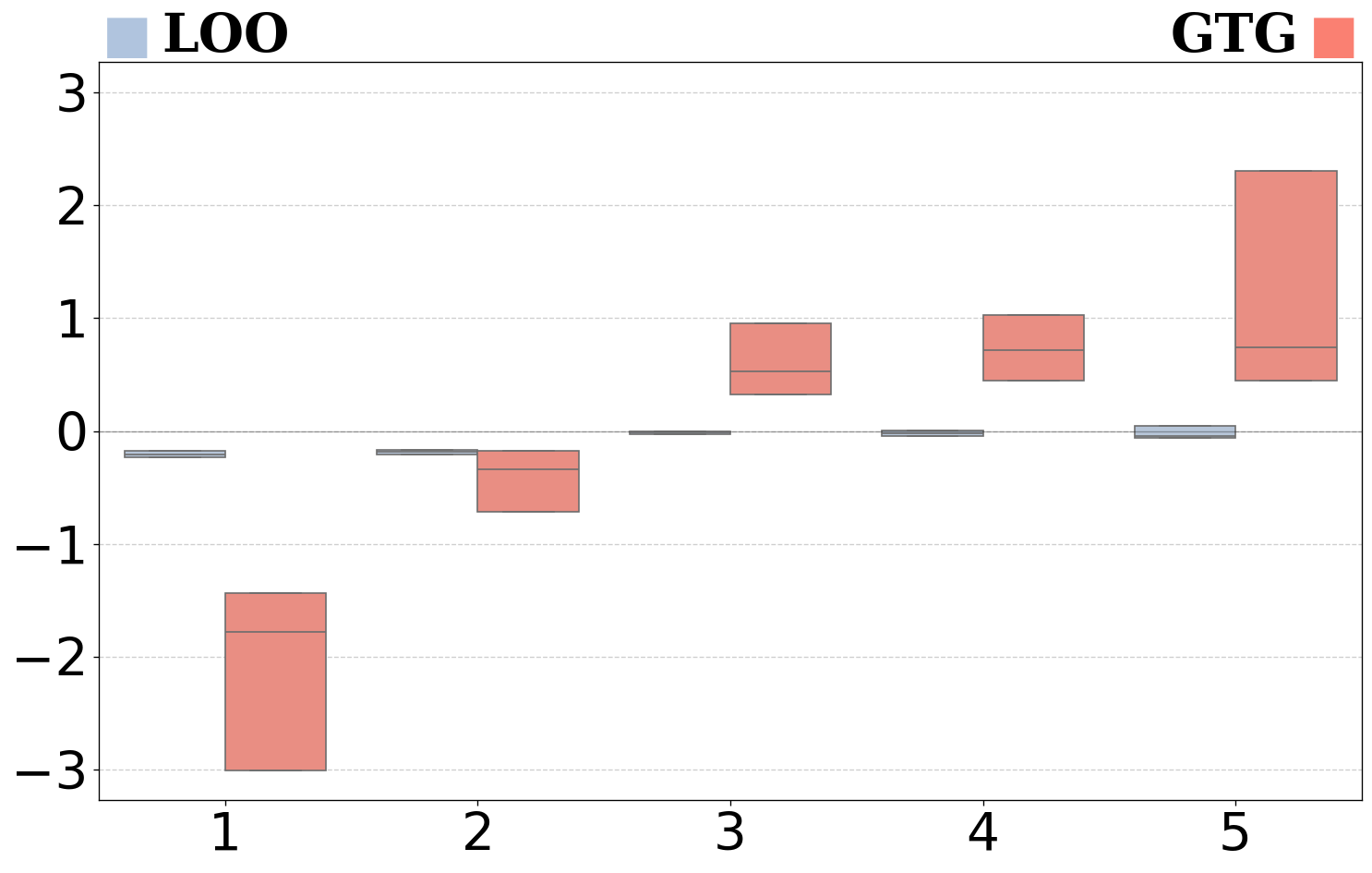}
    \end{subfigure}
    \hfill
    \begin{subfigure}{0.24\textwidth}
        \caption{Fashion/non-IID. }
        \label{fig:TD_diff_C_N}
        \includegraphics[width=\linewidth]{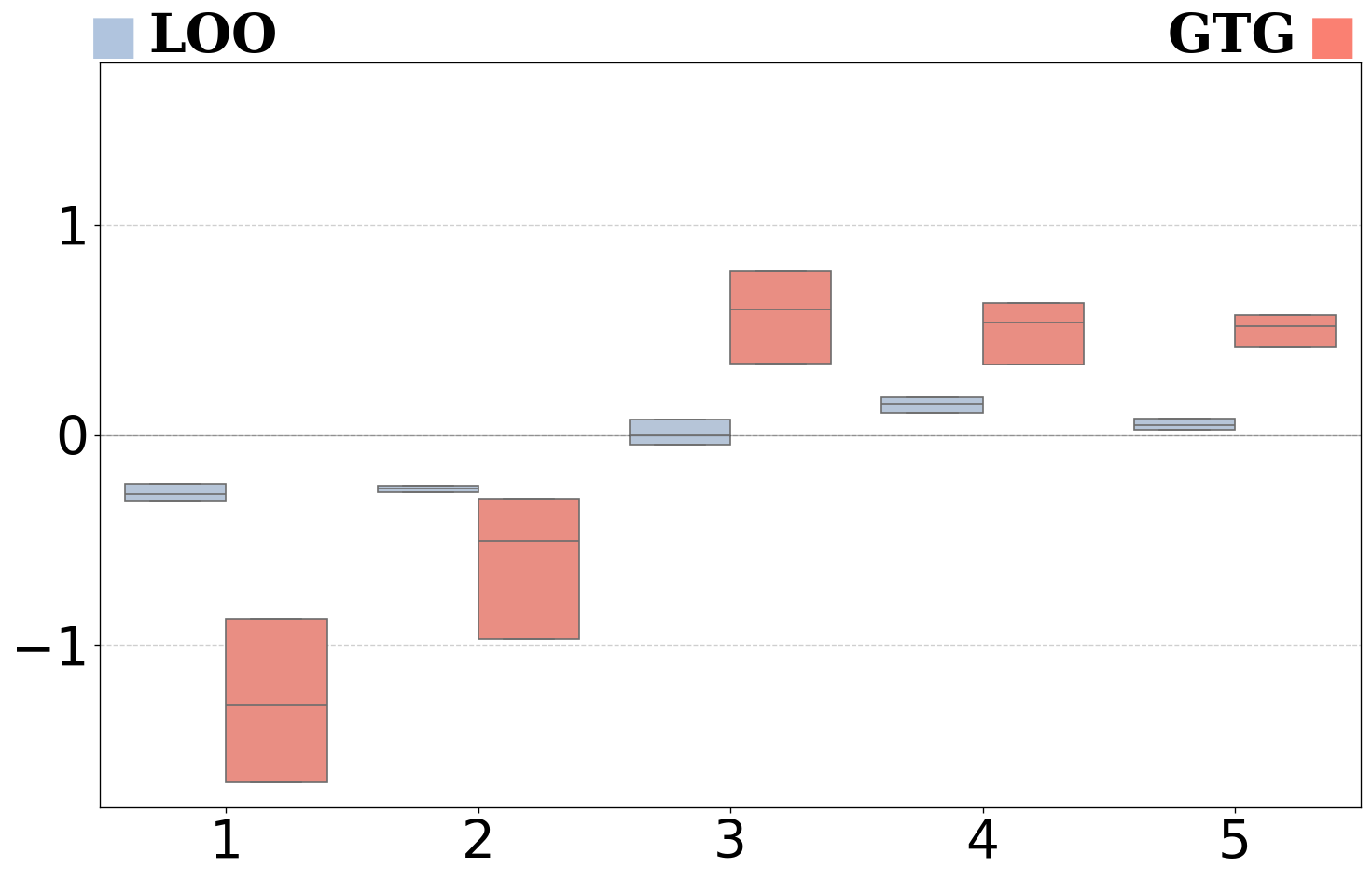}
    \end{subfigure}
\end{figure}

Here, we tested the attacker's (Client 1) ability to decrease a target's score (Client 2) on both the ADULT and FMNIST datasets. Figure~\ref{fig:TD} shows the CE scores for all clients across all cases. This attack seems to be more effective in more complex scenarios: score reduction is more visible on the more complex task (right side) than on the simpler task (left side). Also, the attack is more effective on non-IID data, which is clear when the IID (odd-numbered subfigures) are compared to the non-IID (even-numbered subfigures) splits. This happens because when clients' data are similar, it is difficult to craft an update that harms only one client, selectively; when the target has a unique distribution, they are easier to isolate. A third finding, also visible in Figure~\ref{fig:TD_diff}, is the collateral damage: in all cases, the attacker's own score was reduced more than the target's. Finally, for GTG specifically (bottom subfigures), we observe a clear trend where decreasing the target's score often increases the scores of other non-targeted, benign clients.

The quantitative analysis in Table~\ref{tab:TD_diff} confirms these observations and echoes the pattern from the \emph{Self Improvement} attack: LOO's absolute changes are smaller than GTG's, but its relative changes are much larger. The scale of the impact here is notable, with the attack reducing the victim's score by a significantly greater factor than the inflation attack increased the attacker's (compare to Table \ref{tab:SI_diff}). 

In contrast to \emph{Self Improvement}, where we utilized the entire validation set for the attack, \emph{Target Decrease} only uses a portion of it. This leads to uncertainty in the attack's effectiveness: while unlikely, it could even increase the targets score. Hence, for the statistical test we utilized a two-tailed $t$-tests. The $p$-values are presented in Table~\ref{tab:TD_p}.

The results show that the most resistant setting is ADULT with IID clients, where score differences are statistically insignificant, reinforcing our earlier observation. In contrast, for both ADULT and FMNIST, all non-IID cases exhibit significant score changes. Interestingly, almost all FMNIST cases with IID clients also show statistically significant differences.

\subsection{Defenses}

While it may seem natural to combine the Architectural Sensitivity study with the Intentional Manipulation study, i.e., testing our poisoning attacks under Byzantine fault tolerant aggregations, we deliberately treated these directions separately. Prior work (e.g., the ACE paper~\cite{xu2024ace}) shows that even weaker attacks can ``fly under the radar'' of anomaly detectors, since such defenses are tuned to identify overtly harmful updates rather than subtle manipulations that appear beneficial. Our \emph{Self Improvement} attack falls exactly into this blind spot: as a side product, it increases the performance of the global model while also corrupting contribution scores. Hence, it raises an ethical question: whether it should even be treated as a malicious attack. It manipulates the fairness of the reward allocation scheme of the federation by providing better model updates according to the utilized and commonly agreed performance metric; in fact, it does deserve the elevated score. 

\begin{figure}[t]
\centering
    \caption{The evolution of the loss values when the dataset (ADULT or FMNIST) is split into 5 clients (IID or non-IID) where the 1st client is decreasing the 2nd client's score. GTG is RED, LOO is BLUE, baseline is GREY.}
    \label{fig:loss}
    \centering
    \begin{subfigure}{0.24\textwidth}
        \includegraphics[width=\linewidth]{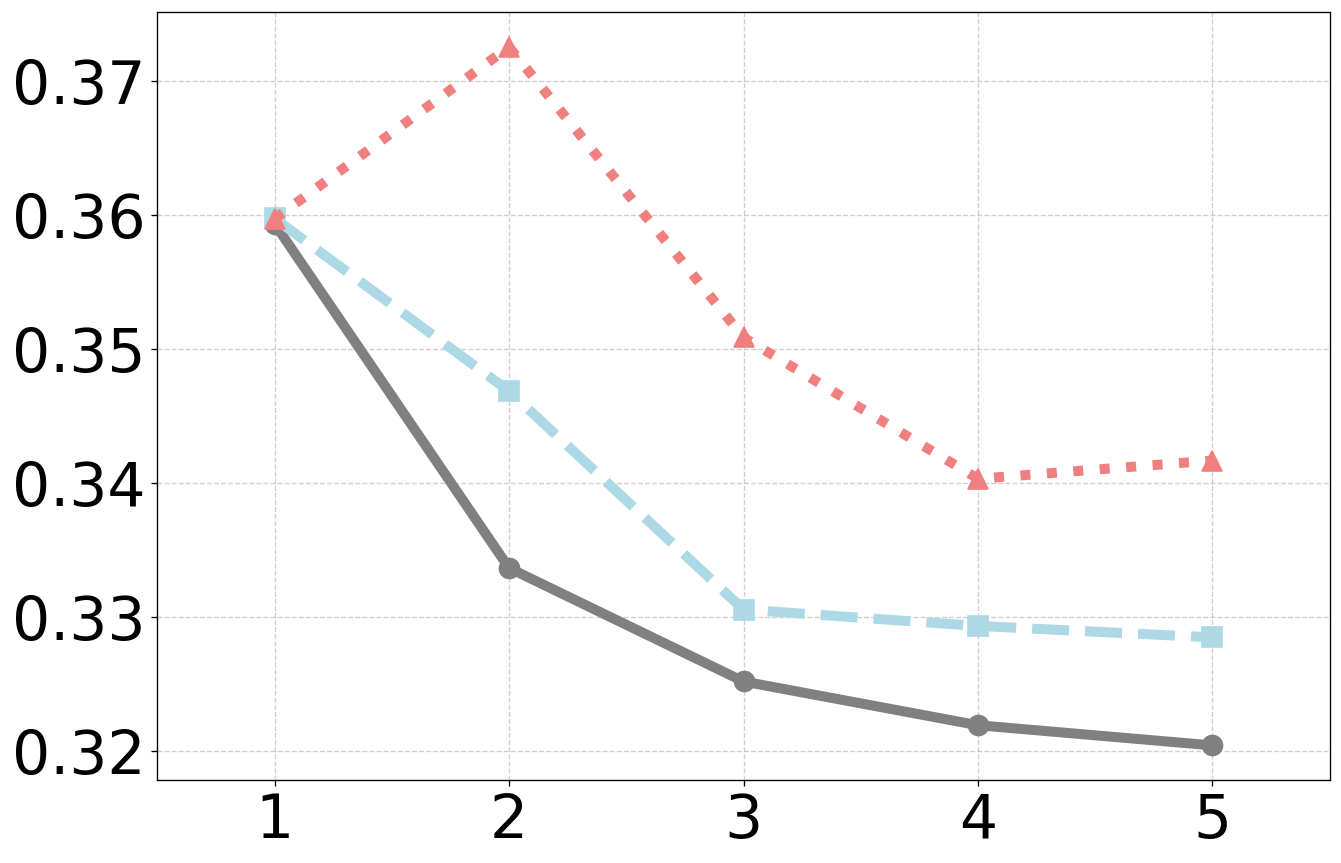}
        \caption{ADULT/IID. }
        \label{fig:targeted_loss_A_I}
    \end{subfigure}
    \hfill
    \begin{subfigure}{0.24\textwidth}
        \includegraphics[width=\linewidth]{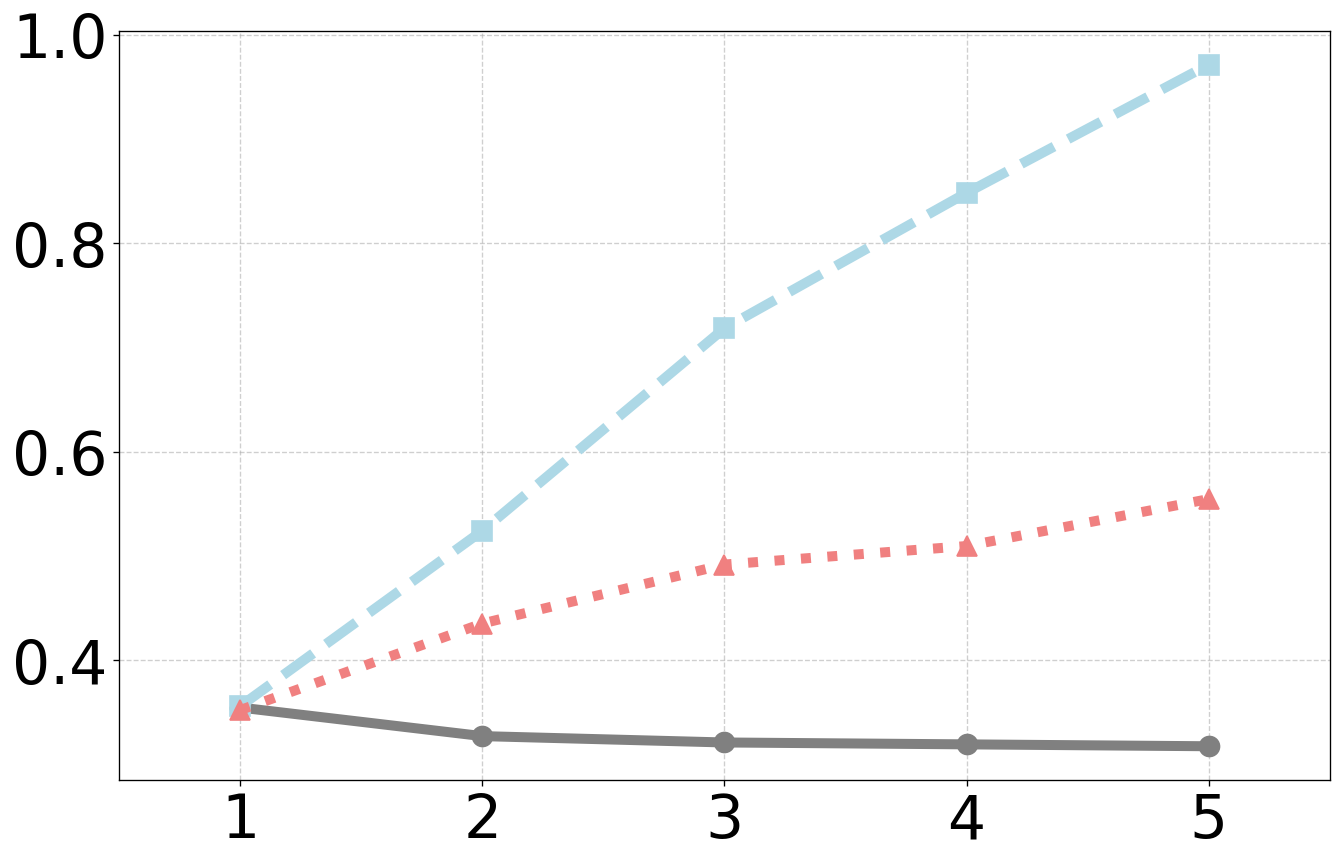}
        \caption{ADULT/non-IID. }
        \label{fig:targeted_loss_A_N}
    \end{subfigure}
    \hfill
    \begin{subfigure}{0.24\textwidth}
        \includegraphics[width=\linewidth]{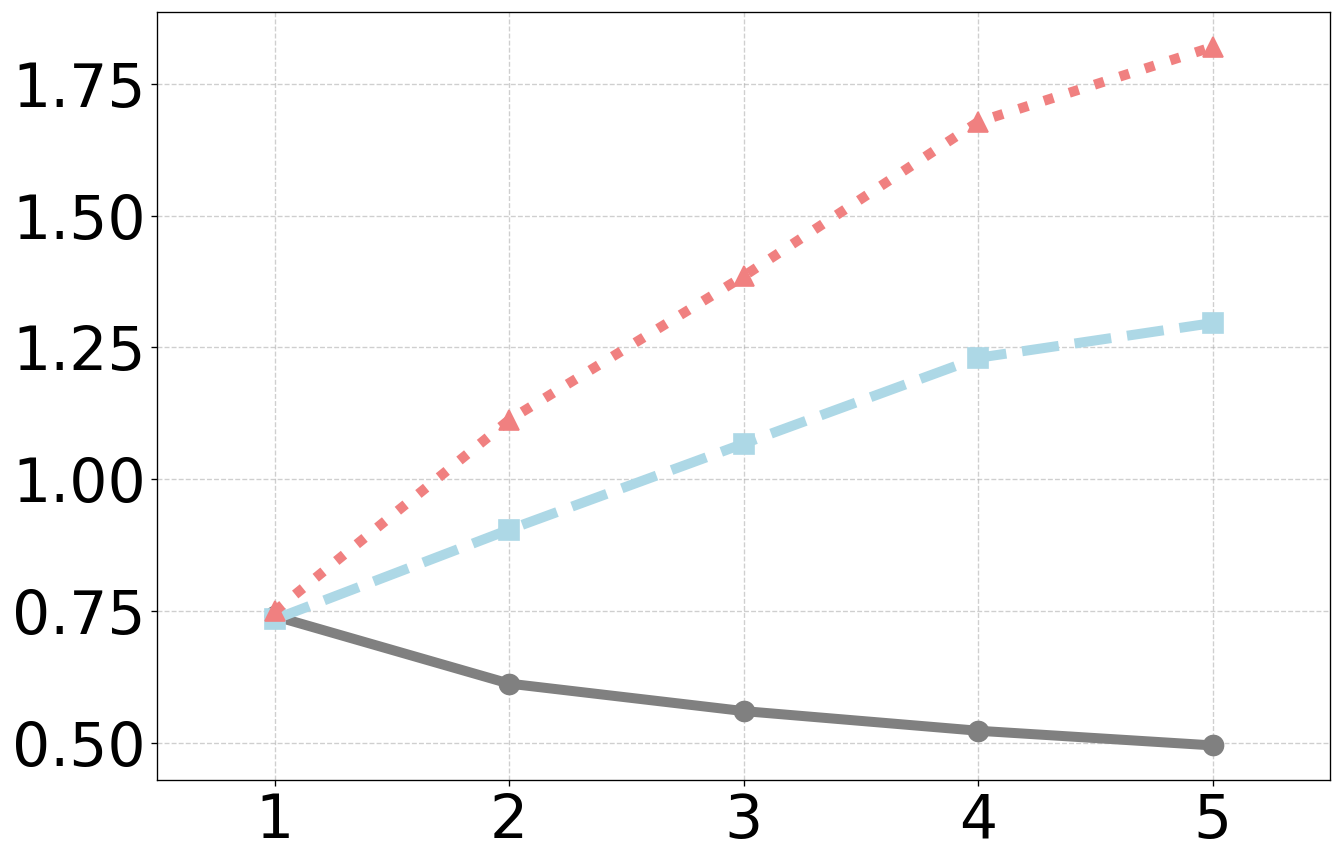}
        \caption{Fashion/IID. }
        \label{fig:targeted_loss_C_I}
    \end{subfigure}
    \hfill
    \begin{subfigure}{0.24\textwidth}
        \includegraphics[width=\linewidth]{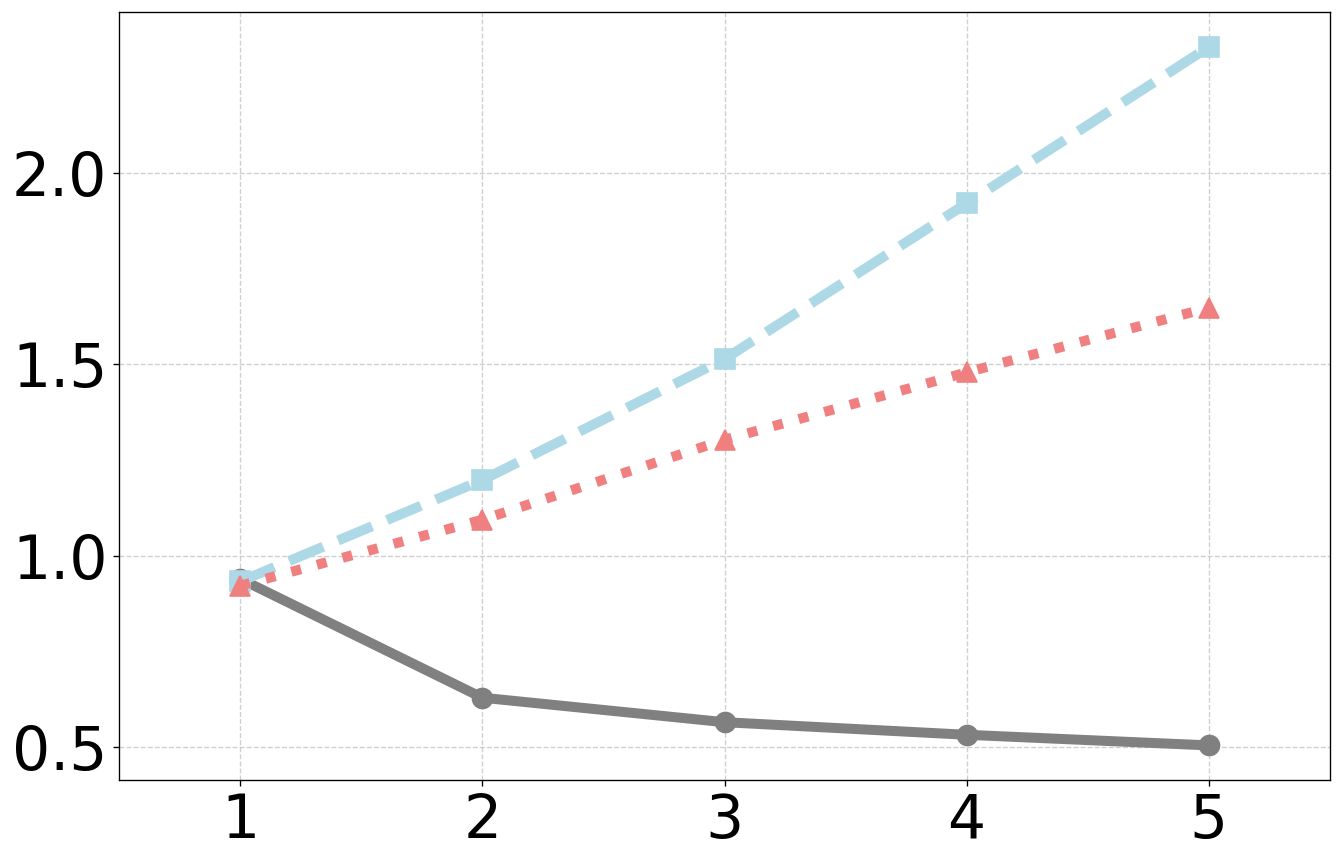}
        \caption{Fashion/non-IID. }
        \label{fig:targeted_loss_C_N}
    \end{subfigure}
\end{figure}

By contrast, defending against the \emph{Targeted Decrease} attack is far more straightforward. Despite our effort to limit the immediate loss change (Equation~\ref{eq:ce_td}), its cumulative effect still leads to significant degradation of the global model over time. As illustrated in Figure~\ref{fig:loss}, repeating the attack over multiple rounds causes the loss to diverge, meaning that simple performance monitoring would be sufficient for detection. 

\subsection{Discussion}

Our two-pronged experimental analysis consistently reveals that contribution scores are highly sensitive to their context. Architecturally, the choice of aggregation method is not a neutral parameter: it directly influences how each client's contribution is evaluated. We observed a clear dichotomy: the distance-based ADP was sensitive to both anomalies and contribution magnitude, making it effective for detecting misbehavior but highly variable across aggregators. In contrast, the marginal-difference-based GTG provides a more stable measure of a client's intrinsic value, but it has a complex and sometimes counter-intuitive interaction with Byzantine fault tolerant mechanisms. 

Adversarially, we show this sensitivity can be exploited in practice: principled, marginal-difference-based CE schemes are directly vulnerable. Although we did not formalize a single utility function, both attacks we presented are reasonable for rational adversaries whose objective is solely to maximize their own score (potentially while preserving model quality) or to suppress a target’s score. Notably, the \emph{Self Improvement} attack is especially insidious---it can improve the global model while corrupting the fairness of the reward system, turning model enhancement into a vulnerability. The \emph{Targeted Decrease} attack, even in its current, not-yet-stealthy form, demonstrates clearly that CE schemes can be manipulated; future work will almost certainly produce more subtle and effective variants. Together, our findings expose a genuine and immediate risk to the integrity of contribution evaluation mechanisms.
\section{Conclusion}
\label{sec:con}

In this work, we demonstrated that contribution evaluation in federated learning is fragile, susceptible to distortions from both well-intentioned architectural choices and adversarial manipulation. 

First, we revealed the impact of the server-side aggregation strategy on client scores. Our experiments showed that employing advanced aggregators to handle client heterogeneity or to provide Byzantine robustness leads to fundamentally different contribution score distributions compared to standard FedAvg. This highlights that a client’s contribution score is not absolute, but depends strongly on the chosen aggregation architecture.

Second, we showed that contribution scores are also vulnerable to direct, malicious attacks. We manipulated marginal difference-based contribution scores, allowing an attacker to inflate their own score. Interestingly, this action often reduced the global loss, blurring the line of what constitutes an attack traditionally. Furthermore, we demonstrated that an informed attacker could effectively reduce a benign target's contribution score.

Ultimately, our dual analysis revealed that a trustworthy contribution evaluation framework requires more than a fair-value, axiom-based scoring scheme like the Shapley Value; it must also be robust to both the implicit biases introduced by the aggregation architecture and the explicit threats posed by strategic adversaries.

\subsection{Limitations}

Our experiments are conducted in a specific cross-silo setting and focus on a representative, yet limited, set of aggregation schemes and CE schemes. Furthermore, our proposed poisoning attacks rely on the strong assumption that the adversary has access to other clients' benign gradients and to the validation dataset. We also prioritized demonstrating the feasibility of score manipulation over its stealthiness, leaving the challenge of untangling model performance from client contribution largely open.

\subsection{Future Work}

Future work should proceed in three key directions. First, the scope of the architectural analysis should be broadened by exploring a wider range of FL scenarios, aggregation methods, and CE schemes. Second, a crucial step is to design more sophisticated attacks that relax strong assumptions, i.e., can the attacks succeed using only a proxy validation dataset, or can the adversary estimate benign gradients using historical data? Finally, future research could address the stealthiness of these attacks by developing optimization strategies with strict constraints to manipulate scores while provably preserving the global model's integrity and performance.

\begin{comment}
\subsection{Acknowledgments}

Project no. 145832, implemented with the support provided by the Ministry of Innovation and Technology from the NRDI Fund, financed under the PD\_23 funding scheme. 
Besides, it is funded by the European Union (Grant Agreement Nr. 101095717, SECURED Project). 
Add OTKA FK, MILAB?
\end{comment}

\section*{LLM usage considerations}

LLMs were used for editorial purposes in this manuscript (none of the authors are native English speakers), and all outputs were inspected by the authors to ensure accuracy and originality. Moreover, we used an LLM to assist with the implementations. Note that the research phase (literature review, identifying research gaps, designing the experiments, evaluating the results) was conducted without utilizing any LLM (or AI in general). 

\bibliographystyle{unsrt}
\bibliography{ref}

\end{document}